%% file: neurips_2026.tex
\documentclass{article}

    \PassOptionsToPackage{numbers, compress}{natbib}

 \usepackage[main, final]{neurips_2026}

\usepackage[utf8]{inputenc} 
\usepackage[T1]{fontenc}    
\usepackage{hyperref}       
\usepackage{url}            
\usepackage{booktabs}       
\usepackage{amsfonts}       
\usepackage{nicefrac}       
\usepackage{microtype}      
\usepackage{xcolor}         

\usepackage{graphicx}

\usepackage{hyperref}

\usepackage{algorithm}      
\usepackage{algpseudocode}  
\usepackage{amsmath}        
\usepackage{amssymb}        
\usepackage{cleveref}
\usepackage{bm}             
\usepackage{enumitem}
\usepackage{pifont} 

\usepackage{tabularx}
\usepackage{caption}
\usepackage{subcaption}

\algnewcommand\LComment[1]{\State \textit{// #1}}

\title{AR-Edit: Training-Free \underline{A}uto\underline{r}egressive Video \underline{Edit}ing without Inversion}

\author{%
  \textbf{Hovhannes Margaryan}$^{1, 3}$ \quad
  \textbf{Quentin Bammey}$^{2}$ \quad
  \textbf{Christian Sandor}$^{1}$ \quad
  \textbf{Vicky Kalogeiton}$^{3}$ \\ \\
  $^{1}$Team ARAI, Universit\'e Paris-Saclay, CNRS, LISN, France \\
  $^{2}$LTCI, T\'el\'ecom Paris, IP Paris, France 
  $^{3}$LIX, \'Ecole Polytechnique, CNRS, IP Paris, France \\[3mm]
  \url{https://marghovo.github.io/AR-Edit/}
}

\usepackage{graphicx}
\usepackage{tabularx}
\usepackage{colortbl}
\usepackage{xcolor}
\usepackage{amssymb}
\usepackage{caption}
\usepackage{subcaption}
\usepackage{makecell}

\usepackage{tikz}
\usetikzlibrary{shapes.geometric}

\usepackage{xcolor}

\usepackage[dvipsnames]{xcolor}

\definecolor{ours}{HTML}{FF7A00}\definecolor{oursedge}{HTML}{A04A00}\definecolor{oursbg}{HTML}{FDEBD9}
\definecolor{mkedge}{HTML}{4D4D4D}\definecolor{mkinv}{HTML}{DBDBDB}\definecolor{mkinvv}{HTML}{A6A6A6}
\definecolor{mkrand}{HTML}{949494}\definecolor{mkfwd}{HTML}{8CACCC}

\definecolor{mknone}{HTML}{A6A6A6}     
\definecolor{mkinvpath}{HTML}{7B52AE}  
\definecolor{mkfwd}{HTML}{4A86C2}      
\definecolor{ours}{HTML}{FF7A00}       

\usepackage[most]{tcolorbox}

\definecolor{findingframe}{HTML}{E07B24}  
\definecolor{findingback}{HTML}{FFF5EB}   
\definecolor{findingtitle}{HTML}{C9651A}  

\newtcolorbox[auto counter]{finding}[1][]{%
  enhanced,
  colback=findingback, colframe=findingframe,
  boxrule=0.8pt, arc=2.5mm, boxsep=0pt,
  left=2.5mm, right=2.5mm, top=1mm, bottom=1.5mm,
  attach boxed title to top left={xshift=3mm, yshift*=-\tcboxedtitleheight/2},
  boxed title style={colback=white, colframe=findingframe, boxrule=0.6pt,
                     arc=1mm, boxsep=0pt, left=1.5mm, right=1.5mm,
                     top=0.6mm, bottom=0.6mm},
  coltitle=findingtitle, fonttitle=\bfseries,
  title={Finding~\thetcbcounter},
  drop fuzzy shadow=black!30,
  before skip=6pt, after skip=6pt,
  #1}

\newcommand{\recV}[1]{\ensuremath{\mathbf{V}^{#1}}}               
\newcommand{\recVset}{\ensuremath{\{\mathbf{V}^{t}\}}}             
\newcommand{\recipe}[3]{\ensuremath{\langle}#1, #2, #3\ensuremath{\rangle}}
\newcommand{\recInv}{\recipe{Inv}{--}{--}}
\newcommand{\recInvInv}{\recipe{Inv}{Inv}{\recVset}}
\newcommand{\recInvFwd}[1][t]{\recipe{Inv}{Fwd}{\recV{#1}}}
\newcommand{\recRandInv}{\recipe{Rand}{Inv}{\recVset}}
\newcommand{\recRandFwd}[1][t]{\recipe{Rand}{Fwd}{\recV{#1}}}
\newcommand{\recOurs}{\recipe{Rand}{Fwd}{\recV{0}}}

\begin{document}

\maketitle

\begin{abstract}

Editing a video stream in real time, without access to future frames, is essential for interactive applications. Existing training-free video editing methods, however, assume full offline access to the entire sequence, while streaming approaches either rely on additional training or are limited to specific editing techniques. We show that this limitation can be removed by analyzing the self-attention dynamics of DMD-distilled autoregressive (AR) video diffusion models. We uncover a key asymmetry: queries and keys remain stable and aligned with their source counterparts across denoising timesteps, while values diverge and encode the evolving content. Inherited from the non-distilled bidirectional teacher, this asymmetry is strongest at high noise, where few-step streaming models operate. The resulting mechanism is simple: a single forward pass suffices to capture source identity in the value features, which can be reused throughout generation. Starting from pure Gaussian noise and injecting only these source values reconstructs the input with fidelity matching or surpassing inversion-based methods, effectively eliminating inversion. Based on this insight, we introduce \textbf{AR-Edit}, a training-free, inversion-free method that caches source values once and selectively injects them during denoising. Across multiple AR backbones, AR-Edit enables real-time streaming editing with superior source preservation and tonal fidelity at minimal overhead.

\end{abstract}

\begin{figure}[!h]
  \centering
  \includegraphics[width=1.0\textwidth]{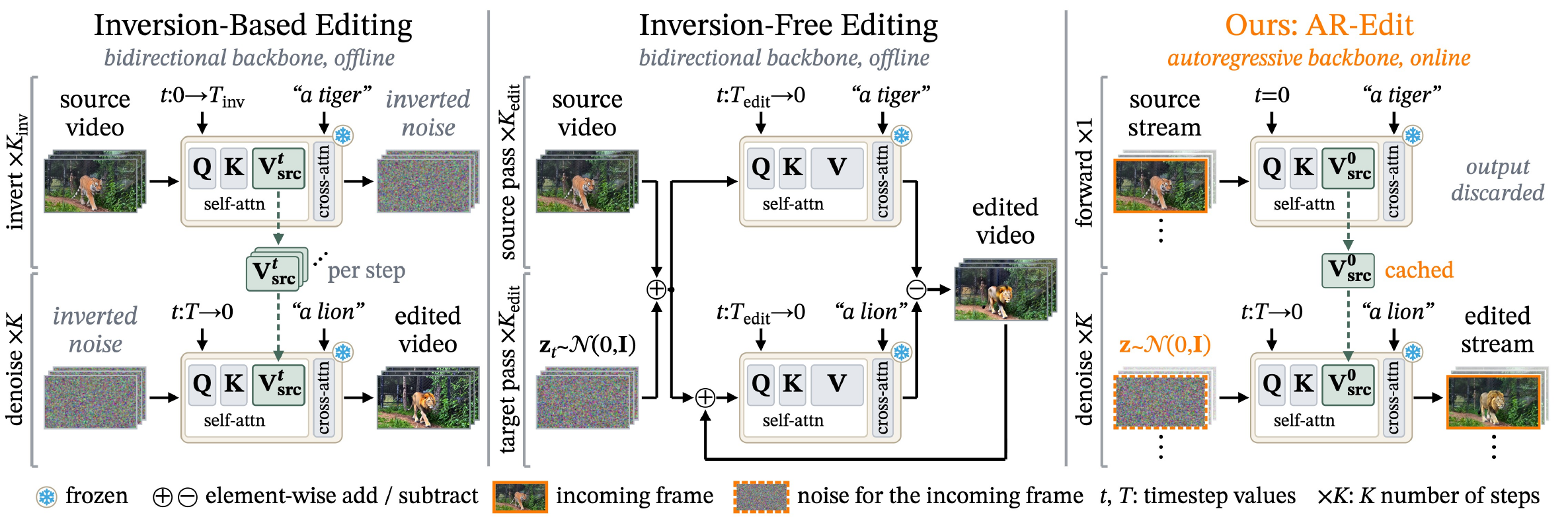}
\caption{\textbf{Training-free video editing paradigms.}
Inversion-based methods~\cite{wang2024taming,deng2024fireflowfastinversionrectified} (left) invert the source to noise and denoise under the target prompt; inversion-free methods~\cite{kulikov2024flowedit,kulikov2026versatileeditingvideocontent} (middle) directly transport the source into the edit. Both need many passes of a bidirectional, offline backbone. \textbf{AR-Edit} (right) runs on an autoregressive backbone~\cite{yin2025causvid,huang2025selfforcingbridgingtraintest,lu2025rewardforcingefficientstreaming}, processing frames causally: it caches source value features in one pass and injects them at denoising pure noise conditioned on the target prompt.\looseness=-1}
  \label{fig:teaser}
\end{figure}

\input{sections/new-intro}
\input{sections/section3_method}
\input{sections/section4_experiments}
\input{sections/section2_related_work}
\input{sections/section5_conclusion}

\input{sections/ack}

\newpage

\bibliographystyle{splncs04}
\bibliography{main}


\newpage
\input{sections/section6_appendix}



\end{document}

%% file: sections/new-intro.tex
\section{Introduction}

Video editing with diffusion models aims to modify a source video according to a target text prompt while preserving everything the prompt does not address: scene layout, identity, background or motion.
Existing training-free methods typically achieve this offline: they require access to the entire video and dozens of model evaluations before producing the first edited frame.
Inversion-based approaches~\cite{wang2024taming,deng2024fireflowfastinversionrectified,Avrahami_2025} map the source to the model's noise space and re-denoise under the target prompt (\Cref{fig:teaser} (left)).
Inversion-free methods such as FlowEdit~\cite{kulikov2024flowedit} and DynaEdit~\cite{kulikov2026versatileeditingvideocontent} avoid this round-trip by constructing a direct transport from the source video to the edited output (\Cref{fig:teaser} (middle)).
In both cases, the editing method requires a bidirectional backbone that sees every frame at once and 20 to 50 network evaluations per edit.

Streaming, which processes frames causally as they arrive, is increasingly relevant for real-time applications such as augmented reality and live content creation. However, existing stream editing methods either rely on additional training \cite{li2025egoeditdatasetrealtimestreaming} or are limited to coarse noise-and-denoise editing \cite{feng2026streamdiffusionv2streamingdynamicinteractive}.

Streaming comes naturally with autoregressive (AR)  video diffusion models. 
CausVid~\cite{yin2025causvid}, Self Forcing~\cite{huang2025selfforcingbridgingtraintest}, and Reward Forcing~\cite{lu2025rewardforcingefficientstreaming} use causal attention to generate a video chunk by chunk without access to future frames.
Each chunk is denoised in only 3 to 4 steps, a regime achieved by distilling a bidirectional teacher through Distribution Matching Distillation (DMD)~\cite{yin2024onestepdiffusiondistributionmatching,yin2024improveddistributionmatchingdistillation}.

Extending the training-free offline editors to the streaming setting remains challenging. Sequential processing rules out aggregation across the full video. Inversion, in turn, requires the many small steps that distilled models lack: with only 3--4 steps, the reverse process no longer recovers the noise that regenerates the source.\looseness=-1

The recent StreamDiffusionV2~\cite{feng2026streamdiffusionv2streamingdynamicinteractive} is the only training-free system that edits on a streaming AR backbone and sidesteps these issues by relying on SDEdit~\cite{meng2022sdeditguidedimagesynthesis}, which adds noise to the source and re-denoises under the target prompt but fails to preserve the source even for small edits.
The concurrent
StreamEdit~\cite{jiao2026streamedittrainingfreevideoediting}\footnote{Released
after our submission.} also targets training-free streaming editing.
However, it requires two network passes per edit step across multiple
denoising steps, exceeding the fixed streaming budget of the backbone.

In this work, we close the gap between offline editing and streaming constraints by turning to self-attention \cite{vaswani2023attentionneed}, which offline inversion-based editors already exploit for source preservation~\cite{wang2024taming,deng2024fireflowfastinversionrectified}.
Examining queries ($\mathbf{Q}$), keys ($\mathbf{K}$), and values ($\mathbf{V}$) in both the DMD-distilled AR and bidirectional settings, we find a clear asymmetry: $\mathbf{Q}$ and $\mathbf{K}$ remain stable across diffusion steps, while $\mathbf{V}$ diverges (\Cref{fig:fig_cs_across_timesteps}).
This property originates in the bidirectional teacher and is strongest at high noise levels, where the few-step AR students exclusively function.

Thus, caching $\mathbf{V}$ from a single clean forward pass of the source and injecting it during denoising from Gaussian noise (\Cref{fig:main_figure} (right)) reconstructs the source with fidelity meeting or surpassing inversion-based methods, while inversion itself fails under the few-step schedule (\Cref{fig:recon}). Inversion is hence redundant in this setting.

Based on these insights, we introduce \textbf{AR-Edit} for streaming video editing (\Cref{fig:teaser} (right)).
For each incoming video chunk, a single forward pass caches $\mathbf{V}$ features, and the edited output is produced by denoising from Gaussian noise under the target prompt while selectively injecting the cached $\mathbf{V}$.

In this process, the target $\mathbf{Q}$ and $\mathbf{K}$ guide the edit semantics, while the injected source $\mathbf{V}$ preserves identity.
In later denoising steps, injection is confined to the unedited region via a mask derived from cross-attention \cite{cai2025ditctrlexploringattentioncontrol,ouyang2025proeditinversionbasededitingprompts}, and the target text tokens are emphasized to strengthen edit adherence \cite{kim2025reflextextguidededitingreal}.
The method operates within the native latency and memory budget of AR models, requires no additional training or optimization, and integrates seamlessly with rolling KV caches~\cite{huang2025selfforcingbridgingtraintest,lu2025rewardforcingefficientstreaming} for streaming.
While trading off some text adherence, AR-Edit improves the quality--speed trade-off, achieving superior source preservation and tonal fidelity on streaming backbones.

Our contributions are as follows:
\begin{itemize}[leftmargin=8mm]
    \item We identify a fundamental asymmetry in the self-attention dynamics of DMD-distilled autoregressive video diffusion models, inherited from their bidirectional teacher: $\mathbf{Q}$ and $\mathbf{K}$ remain stable across diffusion timesteps, while $\mathbf{V}$ diverges and governs the generated content.
    \item We show that injecting $\mathbf{V}$ from a single clean forward pass
    reconstructs the source with fidelity  matching or exceeding inversion-based
    methods, establishing that inversion is unnecessary in this regime.
    \item We propose \textbf{AR-Edit}, a training-free, inversion-free
    method for streaming video editing that caches and injects source
    $\mathbf{V}$ during denoising conditioned on the target prompt.
    \item We introduce a benchmark for evaluating streaming video
    editing under autoregressive constraints and demonstrate that
    AR-Edit improves source preservation and artifact
    suppression while maintaining throughput and latency on streaming backbones.
\end{itemize}

%% file: sections/section3_method.tex
\section{Preliminaries}
\label{sec:Preliminaries}

\subsection{Autoregressive Video Diffusion Models}
\label{sec:prelim_ar}

State-of-the-art text-to-video models such as Wan~2.1~\cite{wan2025wanopenadvancedlargescale} are
\emph{bidirectional}: a Diffusion Transformer (DiT)~\cite{peebles2023scalablediffusionmodelstransformers}
denoises all frames jointly, requiring the entire
sequence---including future frames---to produce a video. This
precludes streaming. Recent work converts these models into
\emph{autoregressive} (AR) generators by replacing bidirectional attention
with a causal one and distilling the result into a few-step
student~\cite{yin2025causvid, huang2025selfforcingbridgingtraintest, liu2025rollingforcingautoregressivelong, lu2025rewardforcingefficientstreaming,  
cui2025selfforcingminutescalehighqualityvideo, kodaira2026streamditrealtimestreamingtexttovideo, yu2025videossmautoregressivelongvideo, hong2025sneakpeekfutureguidedinstructionalstreaming}.

\textbf{Distillation via Distribution Matching Distillation (DMD) ~\cite{yin2024improveddistributionmatchingdistillation,yin2024onestepdiffusiondistributionmatching}.}
Given a pretrained bidirectional teacher that requires ${\sim}50$ denoising
steps, DMD trains a causal student that generates in 3--4 steps by combining a regression loss on teacher-generated pairs with a distribution-matching term
that minimizes an approximate Kullback–Leibler (KL) divergence between the student's output and
the data distribution. This method permits \emph{asymmetric} architectures,  enabling autoregressive generation~\cite{yin2025causvid}.

\begin{figure*}[t]
    \centering
    \includegraphics[width=1.0\linewidth]{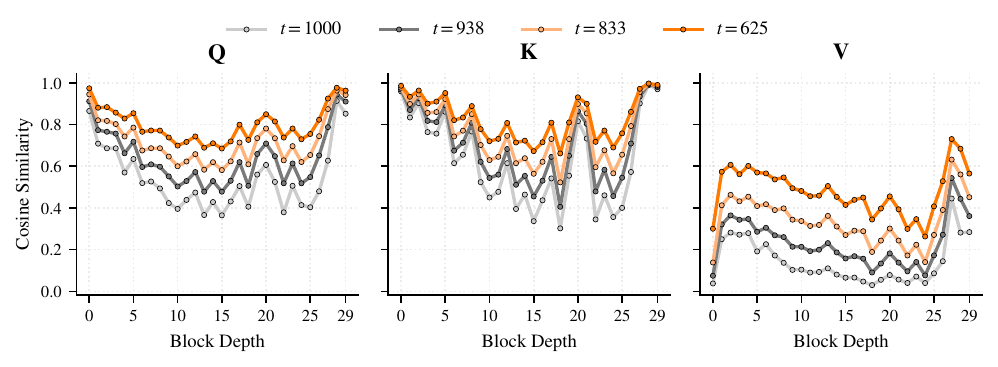}%
    \caption{\textbf{\(\mathbf{Q}\) and \(\mathbf{K}\) are timestep-stable;
\(\mathbf{V}\) is not.} Cosine similarity of
\(\mathbf{Q}\), \(\mathbf{K}\), and \(\mathbf{V}\) to their clean
($t{=}0$) counterparts at each self-attention block of Reward
Forcing~\cite{lu2025rewardforcingefficientstreaming}.
\(\mathbf{Q}\) and \(\mathbf{K}\) retain high similarity
to those obtained from the clean source across all timesteps, while \(\mathbf{V}\) diverges sharply.}

\label{fig:fig_cs_across_timesteps}
\end{figure*}

\textbf{Self-Attention.} Each transformer block in the video DiT \cite{peebles2023scalablediffusionmodelstransformers} computes self-attention \cite{vaswani2023attentionneed} over the
sequence of spatial-temporal tokens. Given an input feature map
$\mathbf{X} \in \mathbb{R}^{N \times d}$, three linear projections produce
queries, keys, and values:
$\mathbf{Q} = \mathbf{X}\mathbf{W}^Q$,\;
$\mathbf{K} = \mathbf{X}\mathbf{W}^K$,\;
$\mathbf{V} = \mathbf{X}\mathbf{W}^V$,
where $\mathbf{W}^Q, \mathbf{W}^K, \mathbf{W}^V \in \mathbb{R}^{d \times d_h}$.
The output is
\begin{equation}
  \text{Attn}(\mathbf{Q},\mathbf{K},\mathbf{V})
  \;=\;
  \text{Softmax}\!\Bigl(\frac{\mathbf{Q}\mathbf{K}^{\top}}{\sqrt{d_h}}
  + \mathbf{M}\Bigr)\,\mathbf{V}\,,
  \label{eq:attn}
\end{equation}
where $\mathbf{M}$ is an attention mask: $\mathbf{M}_{ij}=0$ for all pairs
in bidirectional models, and $\mathbf{M}_{ij}=-\infty$ whenever token~$j$
belongs to a future frame in causal models. The product
$\mathbf{Q}\mathbf{K}^{\top}$ determines \emph{where} each token attends,
while $\mathbf{V}$ determines \emph{what} content is aggregated. This
functional separation is central to the analysis
in~\Cref{sec:Attention_Structure}, where we show that $\mathbf{Q}$ and $\mathbf{K}$
remain stable across different diffusion timesteps while $\mathbf{V}$ diverges sharply (\Cref{fig:fig_cs_across_timesteps}).

\textbf{Inference with KV caching.}
At inference the causal student generates frames causally,
attending to key--value (KV) representations of previously generated frames.
To bound memory, a sliding-window policy retains only the most recent frames,
while a small set of \emph{sink tokens} from the first frames serves as
global anchors throughout the
rollout~\cite{huang2025selfforcingbridgingtraintest,shin2026motionstreamrealtimevideogeneration, lu2025rewardforcingefficientstreaming}.

\subsection{Reconstruction and Editing by Inversion}
\label{sec:prelim_inversion}

A common approach for training-free editing is \emph{inversion}: given a clean
source input $\mathbf{x}_{0}$, i.e.\ the video to be reconstructed or edited, and its prompt $c_{\text{src}}$, one recovers a
noisy latent $\mathbf{z}_{T}$ by integrating the velocity field forward in
time~\cite{song2022denoisingdiffusionimplicitmodels,mokady2022nulltextinversioneditingreal,rout2024semanticimageinversionediting,wang2024taming} that would reproduce the input video if used as the starting point for denoising:
\begin{equation}
  \mathbf{z}_{t+\Delta t}
  \;=\;
  \mathbf{z}_{t}
  \;+\;
  \Delta t\;\boldsymbol{v}_{\theta}(\mathbf{z}_{t},\,t,\,c_{\text{src}})\,,
  \qquad
  \mathbf{z}_{0} = \mathbf{x}_{0}\,,
  \label{eq:inversion}
\end{equation}

where $\boldsymbol{v}_{\theta}$ is the learned velocity, $\mathbf{z}_{t}$ is the latent at noise level $t$, and $\Delta t$ is
the integration step size. Editing then denoises $\mathbf{z}_{T}$ with a target prompt $c_{\text{tar}}$,
injecting source values $\mathbf{V_{\text{src}}}$ from the inversion
trajectory into the target
pass~\cite{wang2024taming,deng2024fireflowfastinversionrectified,Avrahami_2025,ouyang2025proeditinversionbasededitingprompts}
so that target $\mathbf{Q}$ and $\mathbf{K}$ steer \emph{where} to attend
while $\mathbf{V_{\text{src}}}$ preserves \emph{what} content is retrieved.

This pipeline is effective when the integration uses small step sizes and many steps. In the distilled few-step regime, however, the student's schedule uses only 3--4 large steps, violating these assumptions; the inverted $\mathbf{z}_{T}$ lies outside the expected Gaussian noise distribution, making inversion \emph{ineffective} \cite{wu2024turboeditinstanttextbasedimage, deutch2024turboedittextbasedimageediting}. Our analysis in \Cref{sec:Attention_Structure} shows that inversion is also
\emph{unnecessary} in the case of DMD-distilled AR video diffusion models: cached $\mathbf{V}$ from a single source forward pass suffices for both reconstruction and editing, even when starting from random Gaussian noise.

\input{sections/section3_method-v2}

\section{AR-Edit: Autoregressive Video Editing without Inversion}
\label{sec:method}

This section presents \textbf{AR-Edit} (\Cref{fig:teaser} (right)), a simple editing method derived from the two observations above: \textit{cache $\mathbf{V}$ from the source, denoise from random noise with the target prompt, and selectively replace the target's $\mathbf{V}$ with the cached source $\mathbf{V}$.} This is possible because $\mathbf{Q}$ and $\mathbf{K}$ are timestep-invariant while $\mathbf{V}$ is timestep-specific, and $\mathbf{V}$-injection from $t{=}0$ reconstructs the source from random noise as faithfully as inversion. The rest of this section extends this idea to the streaming autoregressive setting, where frames arrive in chunks and each chunk must be edited causally before the next one arrives.

\textbf{Per-chunk $\mathbf{V}$-extraction.}
Consider a stream of source video chunks
$\{\mathbf{x}_{\text{src}}^{(n)}\}_{n=0}^{N-1}$, each containing $F$ frames.
When chunk $n$ arrives, we encode it and run a single forward pass of the
distilled model at $t{=}0$ (the clean endpoint) conditioned on the source
prompt $c_{\text{src}}$. During this network pass we cache the
pre-attention $\mathbf{V}$ tensor at every self-attention block $l > b_{\min}$, where
$b_{\min}$ is a block-depth threshold.  

\textbf{Editing via target-prompted denoising with V-injection.}
To edit chunk $n$, we sample $\mathbf{z} \sim \mathcal{N}(0, \mathbf{I})$ and
denoise it over $K$ steps using the distilled model's predict-then-renoise
schedule \cite{lu2025rewardforcingefficientstreaming, yin2025causvid}, conditioned on the target prompt $c_{\text{tar}}$. At each of the
first $S \leq K$ denoising steps and at every block $l > b_{\min}$, we replace
the target's value with the cached source value:
$\mathbf{V}^{(n,l)} \leftarrow \mathbf{V}_{\text{src}}^{(n,l)}$. The target's $\mathbf{Q}$ and $\mathbf{K}$,
governed by $c_{\text{tar}}$, remain untouched, so the attention routing
reflects the edit: which spatial locations attend to which is determined by the target semantics. The injected source $\mathbf{V}$ controls what content
is read out at each attended location, anchoring preserved regions to the
source appearance.

To localise the edit, we apply the injection with a   \underline{C}ommit-\underline{t}hen-\underline{A}nchor (CtA) schedule. Replacing $\mathbf{V}$ everywhere acts on the whole   frame, whereas most edits are intended for one region, so at the first $S$   steps we inject everywhere and commit to the edit, and from step   $S_{\text{CtA}} \geq S$ onward we inject only outside the edited region,   $\mathbf{V}^{(n,l)} \leftarrow M \odot \mathbf{V}^{(n,l)} + (1-M) \odot \mathbf{V}_{\text{src}}^{(n,l)}$,   where $M \in [0,1]$ is a soft mask derived from the cross-attention of the   tokens that differ between the source and target   prompts~\cite{cai2025ditctrlexploringattentioncontrol,   ouyang2025proeditinversionbasededitingprompts}, read from the blocks   $l \leq b_{\text{CtA}}$ with $b_{\text{CtA}} < b_{\min}$, so the mask is ready   before the injected blocks of the same pass. Any steps between $S$ and   $S_{\text{CtA}}$ run without injection; by default $S_{\text{CtA}} = S$ = 2. This  anchors everything outside $M$ to the source while the edited region stays   free. Additionally, we strengthen the alignment of the generated chunk with the target prompt by a text boost: in every cross-attention layer, the unnormalized attention weights of the target tokens absent from the source prompt are scaled by a constant factor before the softmax normalization (\Cref{app:appendix_imp_details}), following~\cite{kim2025reflextextguidededitingreal}.   \Cref{app:appendix_imp_details} details the mask construction and  text boosting strategies. We ablate $S$, $S_{\text{CtA}}$, $b_{\min}$ and   $b_{\text{CtA}}$ in \Cref{sec:Experiments}.

\textbf{Maintaining causal context across chunks.} In the autoregressive backbone, each chunk attends to past chunks via a rolling
KV cache $\mathcal{K}_{\text{gen}}$. During the editing pass, the keys and
values written into $\mathcal{K}_{\text{gen}}$ at each block are the
\emph{native} target-prompt outputs. This
ensures that when future chunks attend backward, they see representations
consistent with the target prompt rather than a patchwork of source and target
features. $\mathbf{V}$-injection is local to the current chunk's attention computation. Algorithm~\ref{alg:v_inject_editing} in \Cref{app:appendix_imp_details} provides a pseudo-code of our method.

%% file: sections/section3_method-v2.tex
\section{Self-Attention Analysis and Reconstruction} \label{sec:Attention_Structure}

In this section, we analyze how queries ($\mathbf{Q}$), keys ($\mathbf{K}$), and values ($\mathbf{V}$) of self-attention \cite{vaswani2023attentionneed} behave across noise levels in DMD-distilled AR diffusion video models and their bidirectional teacher. We find that $\mathbf{Q}$ and $\mathbf{K}$ remain stable across noise levels while $\mathbf{V}$ diverges, suggesting that the identity of a video can be captured from the value features of a single forward pass. We then validate this through a systematic reconstruction study.

\textbf{Self-attention behavior in DMD-distilled AR and bidirectional models.}\quad
We examine the Wan~2.1 DMD-distilled student from Reward
Forcing~\cite{lu2025rewardforcingefficientstreaming} at each of its
four training timesteps $t \in \{1000, 938, 833, 625\}$ and at $t{=}0$, hereafter the \emph{clean anchor} or
\emph{counterpart}.
For each timestep, we noise a video to~$t$ via linear
interpolation, run a single forward pass, and extract post-RoPE \cite{su2023roformerenhancedtransformerrotary}
$\mathbf{Q}$, post-RoPE $\mathbf{K}$, and pre-attention $\mathbf{V}$
at all 30 self-attention blocks.
We measure the cosine similarity between each feature tensor and
its clean counterpart.

Figure~\ref{fig:fig_cs_across_timesteps} reveals a clear asymmetry.
$\mathbf{Q}$ and $\mathbf{K}$ remain close to the clean anchor across
all timesteps (mean cosine similarity $0.68$ and $0.73$), even at the
highest noise level $t{=}1000$.
$\mathbf{V}$ diverges sharply (mean $0.30$).
In other words, the attention routing is nearly invariant to the
noise level, while the content carried by $\mathbf{V}$ changes
substantially.

\begin{figure*}[t]
    \centering
    \includegraphics[width=1.0\linewidth]{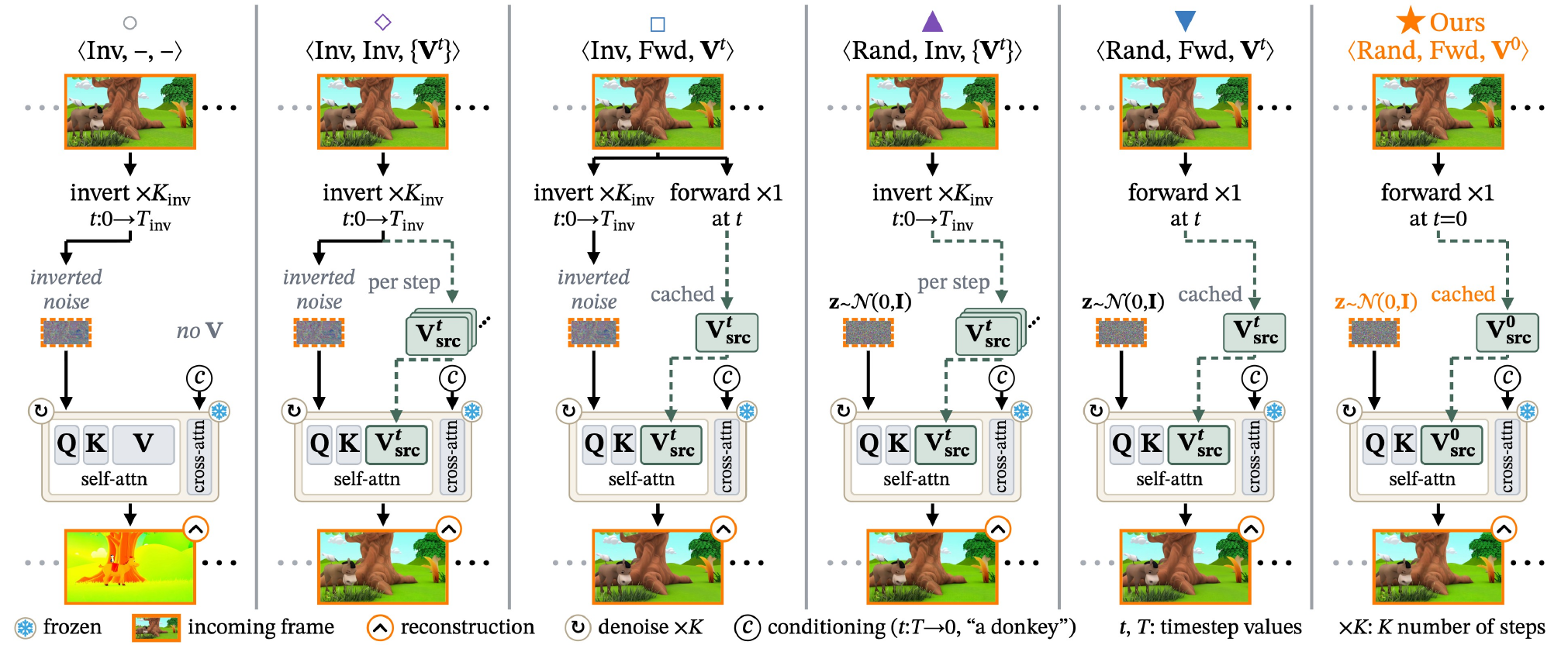}%
\caption{\textbf{Reconstruction Recipes.} Each recipe is named by its initial
noise (Inv: inverted, Rand: random), $\mathbf{V}$ source (Inv: inversion,
Fwd: forward pass), and $\mathbf{V}$ timestep ($\{\mathbf{V}^{t}\}$: one
$\mathbf{V}$ per denoising step, $\mathbf{V}^{t}$: a single $\mathbf{V}$ at a
fixed $t$). Ours \emph{\recOurs} removes inversion entirely by
starting from random noise with $\mathbf{V}$ from a clean forward pass at $t{=}0$.}
    \label{fig:main_figure}
\end{figure*}

To trace the origin of this asymmetry, we repeat the analysis on the non-distilled bidirectional Wan~2.1
1.3B teacher~\cite{wan2025wanopenadvancedlargescale} across the full
noise range (Appendix~\ref{app:qkv_models},
Figure~\ref{fig:qkv_teacher}).
The teacher exhibits the same asymmetry, and its strength depends on
the noise level: at high noise, $\mathbf{Q}$ and $\mathbf{K}$ stay
near the clean anchor while $\mathbf{V}$ departs from it; toward low
noise the gap narrows.
The asymmetry is thus a property of the pretrained model that the
student inherits through distillation.
Crucially, it is strongest where the student operates: its training
timesteps satisfy $t \geq 625$, placing them in the high-noise
regime, and sampling never reaches the low-noise region where the
asymmetry fades.

Appendix~\ref{app:qkv_extended} generalizes this analysis to
additional DMD-distilled students, both autoregressive and
bidirectional~\cite{yin2025causvid}, and replicates it during
generation from noise and during editing
(\Cref{alg:v_inject_editing}).

\begin{finding}
In DMD-distilled autoregressive video
diffusion models and their pretrained bidirectional teacher, self-attention
$\mathbf{Q}$ and $\mathbf{K}$ remain stable across noise levels
while $\mathbf{V}$ diverges from the clean anchor. The asymmetry
originates in the teacher and is strongest in the high-noise regime
where the distilled students operate.
\end{finding}

Since $\mathbf{V}$ carries the video identity while $\mathbf{Q}$
and $\mathbf{K}$ remain structurally compatible with the clean anchor
across noise levels, a single forward pass of the video at $t{=}0$
suffices to capture its identity in the value features.
These cached $\mathbf{V}$ features can then be injected during
denoising with the target prompt, without disrupting the attention
routing governed by $\mathbf{Q}$ and $\mathbf{K}$.
We validate this strategy in the reconstruction setting next.

\textbf{Source Reconstruction.}\quad
Existing training-free editors rely on inversion
(\Cref{sec:prelim_inversion}) to recover a noise tensor $z_T$ that
anchors the edit to the source.
If $\mathbf{V}$ from $t{=}0$ already carries the source identity in
the DMD-distilled AR setting, this step may be
unnecessary. We test this with six reconstruction recipes (\Cref{fig:main_figure}) on our benchmark, defined by three choices: the initial noise, where the injected $\mathbf{V}$ tensors come from, and the timestep at which
they are taken.
Each recipe is named after these three choices: the initial noise is inverted (Inv) or random (Rand); $\mathbf{V}$ comes from the inversion path (Inv) or a single forward pass (Fwd), with -- when no $\mathbf{V}$ is injected; and \recVset{} denotes one $\mathbf{V}$ per denoising step, while \recV{t} is a single $\mathbf{V}$ taken at a fixed $t$.

In baseline \emph{\recInv}, the clean source is inverted to obtain a noise tensor $z_T$, from which denoising is done without $\mathbf{V}$ injection. In \emph{\recInvInv}, the $\mathbf{V}$ tensors captured during inversion are replayed at each corresponding denoising step, matching each $\mathbf{V}$ to its timestep. In \emph{\recInvFwd}, $z_T$ is still obtained by inversion, but $\mathbf{V}$ comes from a single forward pass: the source is noised to a fixed timestep $t$, passed through the network, and the resulting $\mathbf{V}$ is injected at every denoising step. In \emph{\recRandInv}, $z$ is drawn from $\mathcal{N}(0,\mathbf{I})$, while $\mathbf{V}$ is still taken from the inversion path at each timestep $t$. In \emph{\recRandFwd}, $z$ is drawn from $\mathcal{N}(0,\mathbf{I})$ and $\mathbf{V}$ comes from a single forward pass of the source noised to a fixed $t$. Finally, in \emph{Ours: \recOurs}, we start from random noise and inject $\mathbf{V}$ cached from a single clean forward pass at $t{=}0$.\looseness=-1

\begin{figure*}[!t]
  \centering
  \includegraphics[width=\linewidth]{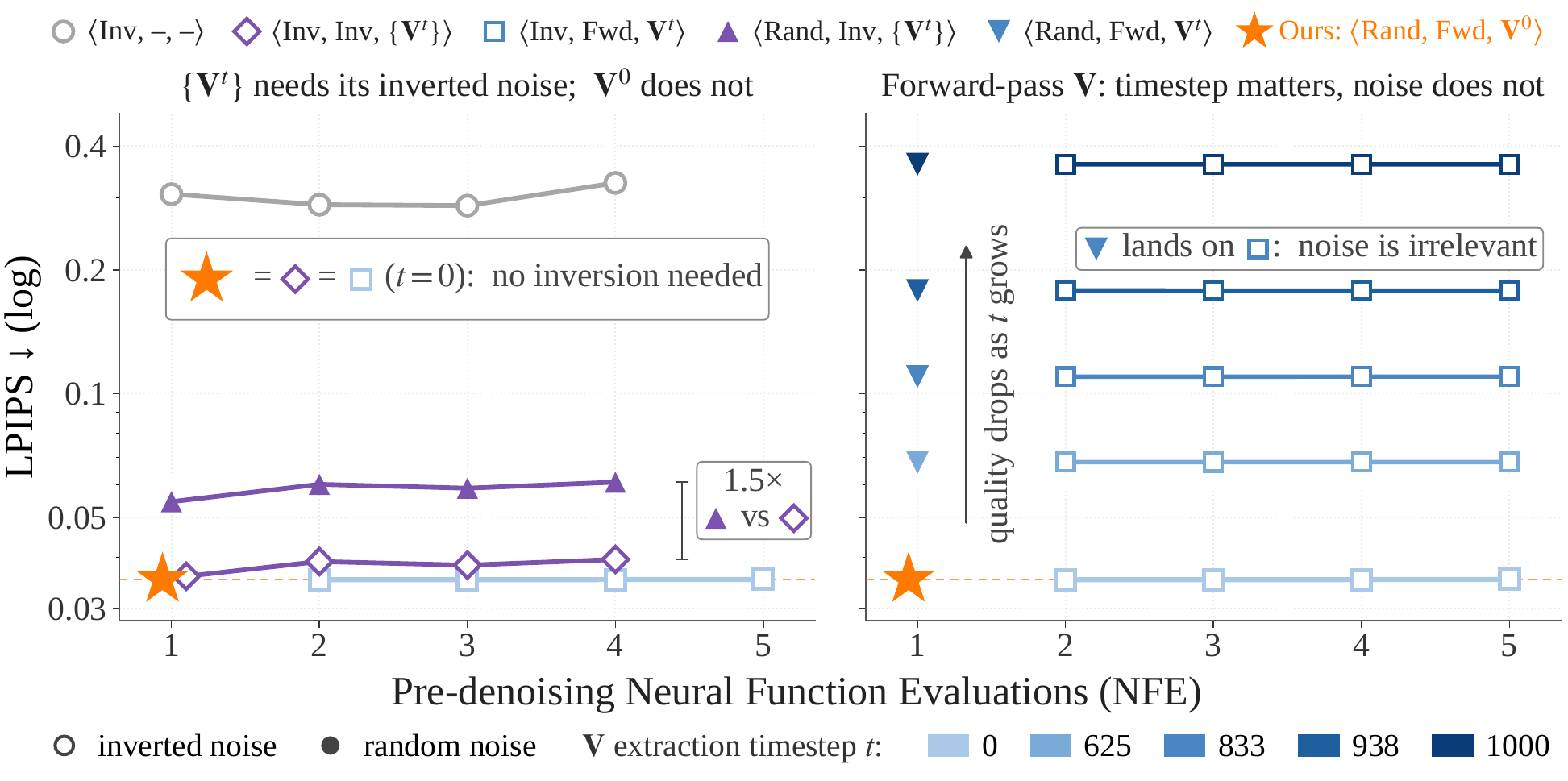}
  \caption{
  \textbf{LPIPS \cite{zhang2018unreasonableeffectivenessdeepfeatures} as a function of the number of pre-denoising NFEs.} Ours performs similarly or surpasses all other reconstruction recipes with one pre-denoising NFE. \looseness=-1}
  \label{fig:recon}
\end{figure*}

We observe that pure inversion collapses at every pre-denoising
neural function evaluation (NFE) budget (\Cref{fig:recon} (left)).
Here, pre-denoising NFE counts the network passes spent to obtain
the initial noise and $\mathbf{V}$ before denoising begins.
\emph{\recInvInv} comes close to ours but never
exceeds it, and only with inverted $z_T$: replaying the same
per-step $\mathbf{V}$ from random noise (\emph{\recRandInv}) raises LPIPS ${\sim}1.5{\times}$ at every
NFE, so per-step $\mathbf{V}$ is tied to the trajectory it was
captured on.
With $\mathbf{V}$ from a single clean pass ($t{=}0$) the initial
noise no longer matters: \emph{\recInvFwd[0]} and ours coincide.
For the fixed-$t$ recipes, $t$ alone sets the quality: it degrades
as $t$ grows and is identical for inverted and random initial noise
(\Cref{fig:recon} (right)).
See visuals in \Cref{app:appendix_additional_eval}.\looseness=-1

\begin{finding}
Drawing the initial noise from
$\mathcal{N}(0,\mathbf{I})$ and injecting $\mathbf{V}$ from $t{=}0$
obtained via a single forward pass of the source video (Ours)
matches or outperforms every more expensive variant at a single
pre-denoising NFE with neither inversion nor per-step matching of
$\mathbf{V}$. Therefore, inversion is not only ineffective in the DMD-distilled AR setting but also unnecessary.
\end{finding}

%% file: sections/section4_experiments.tex
\section{Experiments}
\label{sec:Experiments}

\begin{table*}[t]
  \centering
  \footnotesize
  \caption{\textbf{Quantitative comparison with the state of the art on the benchmark.} Best in \textbf{bold}, second-best \underline{underlined}, per backbone block. TF measures preservation of the source intensity distribution (higher = fewer artifacts). Peak Mem.\ reports allocated / reserved GPU memory.} 
  \label{tab:main_comparison}
  \setlength{\tabcolsep}{1.8pt}
  \renewcommand{\arraystretch}{1.12}
  \begin{tabular}{@{}l<{\hspace{7pt}} cc >{\hspace{7pt}}cccc >{\hspace{7pt}}ccccc@{}}   
    \toprule
    & \multicolumn{2}{c}{\textbf{Editing Quality}} & \multicolumn{4}{>{\hspace{7pt}}c}{\textbf{Overall Visual Fidelity}} & \multicolumn{5}{>{\hspace{7pt}}c}{\textbf{Runtime \& Memory}} \\
    \cmidrule(lr){2-3} \cmidrule(l{9pt}r){4-7} \cmidrule(l{9pt}r){8-12}
    Method
      & \makecell[b]{Source\\Adh.\,$\uparrow$} & \makecell[b]{Text\\Adh.\,$\uparrow$}
      & \makecell[b]{Video\\Quality\,$\uparrow$} & \makecell[b]{CLIP\\Frame\,$\uparrow$} & TF\,$\uparrow$ & TF$_{\text{u}}$\,$\uparrow$
      & \makecell[b]{Edit\\Steps} & NFE\,$\downarrow$ & \makecell[b]{Throughput\\(FPS)\,$\uparrow$} & \makecell[b]{Latency\\(s)\,$\downarrow$} & \makecell[b]{Peak Mem.\\(GB)\,$\downarrow$} \\
    \midrule
    \multicolumn{12}{@{}l}{\emph{Bidirectional backbone (Wan 2.1 1.3B, non-distilled \cite{wan2025wanopenadvancedlargescale})}} \\
    \addlinespace[3pt]
    ~~~SDEdit \cite{meng2022sdeditguidedimagesynthesis}  & 5.01 & 6.12 & \textbf{7.98} & 0.84 & 0.39 & 0.31 & 45 & \textbf{45} & \textbf{0.85} & \textbf{94.9} & \textbf{24.8 / 28.3} \\
    ~~~FlowEdit \cite{kulikov2024flowedit} & \textbf{7.12} & \underline{6.22} & \underline{7.86} & \underline{0.88} & \textbf{0.49} & \underline{0.44} & 45 & \underline{90} & 0.45 & 181.8 & \underline{24.9 / 31.1} \\
    ~~~DynaEdit \cite{kulikov2026versatileeditingvideocontent} & \underline{6.97} & \textbf{6.26} & 7.80 & \underline{0.88} & \underline{0.48} & \textbf{0.45} & 45 & 94 & 0.42 & 195.1 & 25.0 / 31.1 \\
    ~~~\emph{\recInvInv}   & 6.89 & 5.13 & 7.77 & \textbf{0.89} & 0.42 & 0.35 & 45 & \underline{90} & \underline{0.52} & \underline{155.5} & 27.8 / 30.3 \\   
    \midrule
    \multicolumn{12}{@{}l}{\emph{Autoregressive (AR) backbone --- Reward Forcing, distilled \cite{lu2025rewardforcingefficientstreaming}}} \\
    \addlinespace[3pt]
    ~~~SDEdit \cite{meng2022sdeditguidedimagesynthesis}        & 3.81 & \textbf{6.54} & 7.92 & 0.81 & 0.27 & 0.23 & 4 & \textbf{5} & \textbf{11.20} & \textbf{0.91} & \textbf{25.7 / 28.9} \\
    ~~~FlowEdit \cite{kulikov2024flowedit}      & 6.77 & 5.98 & 7.89 & \underline{0.89} & 0.38 & 0.34 & 4 & 9 & 6.72 & \underline{1.22} & 30.8 / 34.2 \\
    ~~~DynaEdit  \cite{kulikov2026versatileeditingvideocontent}     & \underline{7.17} & 5.96 & 7.85 & \textbf{0.89} & \underline{0.39} & \underline{0.36} & 4 & 13 & 4.66 & 1.63 & 30.8 / 34.2 \\
    ~~~\emph{\recInvInv} & 4.93 & 5.85 & \underline{7.93} & 0.85 & 0.38 & 0.32 & 4 & 9 & 9.07 & 1.49 & 27.3 / 31.0 \\
    \rowcolor{oursbg}
    ~~~\textcolor{ours}{\textbf{Ours}} & \textbf{7.28} & \underline{6.11} & \textbf{7.96} & 0.89 & \textbf{0.47} & \textbf{0.41} & 4 & \underline{6} & \underline{10.50} & \underline{1.22} & \underline{27.2 / 30.5} \\
    \midrule
    \multicolumn{12}{@{}l}{\emph{Autoregressive (AR) backbone --- CausVid, distilled \cite{yin2025causvid}}} \\
    \addlinespace[3pt]
    ~~~SDEdit  \cite{meng2022sdeditguidedimagesynthesis}       & 2.78 & 6.45 & \textbf{8.15} & 0.79 & 0.25 & 0.20 & 3 & \textbf{4} & \textbf{15.30} & \textbf{0.91} & \textbf{29.5 / 34.0} \\
    ~~~FlowEdit  \cite{kulikov2024flowedit}      & 3.98 & \textbf{6.88} & 7.70 & 0.81 & 0.36 & 0.28 & 3 & 7 & 11.30 & 1.37 & 35.2 / 39.6 \\
    ~~~DynaEdit  \cite{kulikov2026versatileeditingvideocontent}      & \underline{4.83} & \underline{6.82} & 7.67 & \underline{0.83} & \underline{0.38} & \underline{0.31} & 3 & 11 & 8.08 & 1.73 & 35.2 / 39.6 \\
    ~~~\emph{\recInvInv} & 3.06 & 6.42 & \underline{8.12} & 0.81 & 0.29 & 0.23 & 3 & 7 & 11.16 & 1.38 & \underline{31.1 / 35.4} \\
    \rowcolor{oursbg}
    ~~~\textcolor{ours}{\textbf{Ours}} & \textbf{6.99} & 6.67 & 8.01 & \textbf{0.84} & \textbf{0.42} & \textbf{0.39} & 3 & \underline{5} & \underline{11.91} & \underline{1.27} & \underline{31.1 / 35.4} \\
    \midrule
    \multicolumn{12}{@{}l}{\emph{Autoregressive (AR) backbone --- Self Forcing, distilled \cite{huang2025selfforcingbridgingtraintest}}} \\
    \addlinespace[3pt]
    ~~~\textcolor{gray}{StreamEdit \cite{jiao2026streamedittrainingfreevideoediting}} & \textcolor{gray}{8.11} & \textcolor{gray}{5.56} & \textcolor{gray}{7.91} & \textcolor{gray}{0.91} & \textcolor{gray}{0.42} & \textcolor{gray}{0.39} & \textcolor{gray}{15} & \textcolor{gray}{32} & \textcolor{gray}{2.27} & \textcolor{gray}{3.22} & \textcolor{gray}{41.3 / 43.4} \\
    ~~~StreamEdit \cite{jiao2026streamedittrainingfreevideoediting} & \underline{7.17} & \textbf{5.87} & \underline{7.56} & \underline{0.88} & \underline{0.42} & \underline{0.38} & 4 & \underline{10} & \underline{6.12} & \underline{1.38} & \underline{41.3 / 42.8} \\
    \rowcolor{oursbg}
    ~~~\textcolor{ours}{\textbf{Ours}} & \textbf{7.22} & \underline{5.77} & \textbf{7.91} & \textbf{0.89} & \textbf{0.46} & \textbf{0.42} & 4 & \textbf{6} & \textbf{10.25} & \textbf{0.85} & \textbf{31.8 / 35.0} \\
    \bottomrule
  \end{tabular}
\end{table*}

\textbf{Dataset.} To the best of our knowledge, no public benchmark exists for evaluating general editing on autoregressive video diffusion models. We build one from $120$ hand-curated Pexels~\cite{pexels} videos: 10 editing categories of $12$ clips each, inspired by ~\cite{kulikov2026versatileeditingvideocontent, ju2023direct}. Each clip has a source caption, drafted by Qwen2.5-VL~\cite{bai2025qwen25vltechnicalreport} and refined by the authors, and a hand-authored target prompt. More in \Cref{app:benchmark}.

\textbf{Evaluation Metrics}. We evaluate editing and visual quality, and speed using both VLM-based and
classical metrics. For \emph{editing}, we follow the VLM-as-judge protocol of
DynaEdit~\cite{kulikov2026versatileeditingvideocontent,hsu2024autovfxphysicallyrealisticvideo,yatim2025dynvfxaugmentingrealvideos,garibi2025tokenverseversatilemulticonceptpersonalization,peng2025dreambenchhumanalignedbenchmarkpersonalized}:
\emph{source adherence} scores how faithfully unedited elements are preserved
(judge sees both videos and the target prompt), while \emph{text adherence}
scores how completely the edit realizes the target prompt (judge sees only
the edited clip and text). Both use a 1--10 integer scale with
Qwen2.5-VL-72B-Instruct~\cite{bai2025qwen25vltechnicalreport} as the judge.
For \emph{visual quality}, we report VLM-judged \emph{video quality} (1--10,
without access to source or prompt) and \emph{CLIP
Frame}~\cite{radford2021learningtransferablevisualmodels,kong2025tamingflowbasedi2vmodels, li2025flowdirectortrainingfreeflowsteering}
(mean pairwise similarity of consecutive frames).  \looseness=-1

\begin{figure*}[ht]
    \centering
    \includegraphics[width=1.0\linewidth]{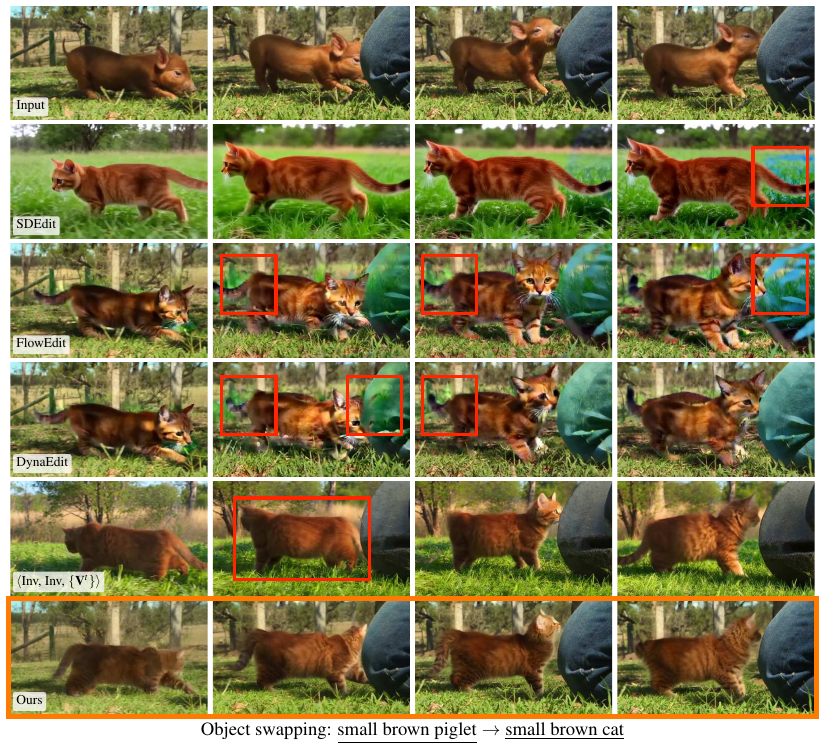}
\caption{\textbf{Visual comparison of editing.}
Red boxes highlight burning artifacts and identity loss.
Ours preserves source appearance while applying the
target edit without tonal artifacts.}

\label{fig:visual_comparison_with_sota}
\end{figure*}

To quantify burning artifacts observed in baseline results
(\Cref{fig:visual_comparison_with_sota}), we propose \emph{Tonal Fidelity} (TF):

\begin{equation}
  \text{Tonal Fidelity} = \frac{1}{1 + D_{\mathrm{KL}}}, \quad
  D_{\mathrm{KL}} = \frac{1}{F}\sum_{f=1}^{F}
  D_{\mathrm{KL}}\!\bigl(h^{\mathrm{edit}}(f) \,\|\, h^{\mathrm{src}}(f)\bigr),
  \label{eq:tonal_fidelity}
\end{equation}
where $h^{\mathrm{edit}}(f)$ and $h^{\mathrm{src}}(f)$ are the normalized 50-bin histograms of the HSV value channel over the highlight pixels ($V \geq 0.9$) of frame $f$, where burning and overexposure manifest, and $F$ is the total number of frames. Values near $1$ indicate a preserved tonal distribution; low values signal overexposure and burning artifacts. We additionally report \emph{Tonal Fidelity unedited (TF$_{\text{u}}$)}, which restricts the metric to the region the edit is not supposed to change: given a per-clip mask of the subject, the histograms of Eq.~\eqref{eq:tonal_fidelity} are computed on the complement of the mask.


For \emph{efficiency}, we report editing steps, NFE per chunk, throughput (FPS), time-to-first-frame latency, and peak GPU memory (allocated~/~reserved, GB) on a single H100. NFE counts all forward passes per edited chunk: velocity-difference editors (FlowEdit, DynaEdit) use two per edit step, and AR backbones add one cache-refresh pass for all methods (Step~3, \Cref{alg:v_inject_editing}). For the bidirectional backbone, reported NFEs exclude classifier-free guidance~\cite{ho2022classifierfreediffusionguidance} overhead.\looseness=-1

\textbf{Implementation.} We implement AR-Edit on three DMD-distilled causal students of Wan~2.1~\cite{wan2025wanopenadvancedlargescale}: Reward Forcing~\cite{lu2025rewardforcingefficientstreaming} and Self Forcing~\cite{huang2025selfforcingbridgingtraintest} (both use a 4-step denoising schedule) and CausVid~\cite{yin2025causvid} (3-step schedule). All compared methods---SDEdit~\cite{meng2022sdeditguidedimagesynthesis}, FlowEdit~\cite{kulikov2024flowedit}, DynaEdit~\cite{kulikov2026versatileeditingvideocontent}, and our baseline \emph{\recInvInv}---are adapted to run autoregressively on the same backbones. We additionally report them in their native setting on bidirectional Wan~2.1~1.3B. We further compare with the concurrent StreamEdit~\cite{jiao2026streamedittrainingfreevideoediting} on Self Forcing. Details in \Cref{app:appendix_imp_details}.

\begin{figure}[t]
  \centering
  \begin{minipage}[t]{0.48\linewidth}
    \vspace{0pt}
    \centering
    \includegraphics[width=\linewidth]{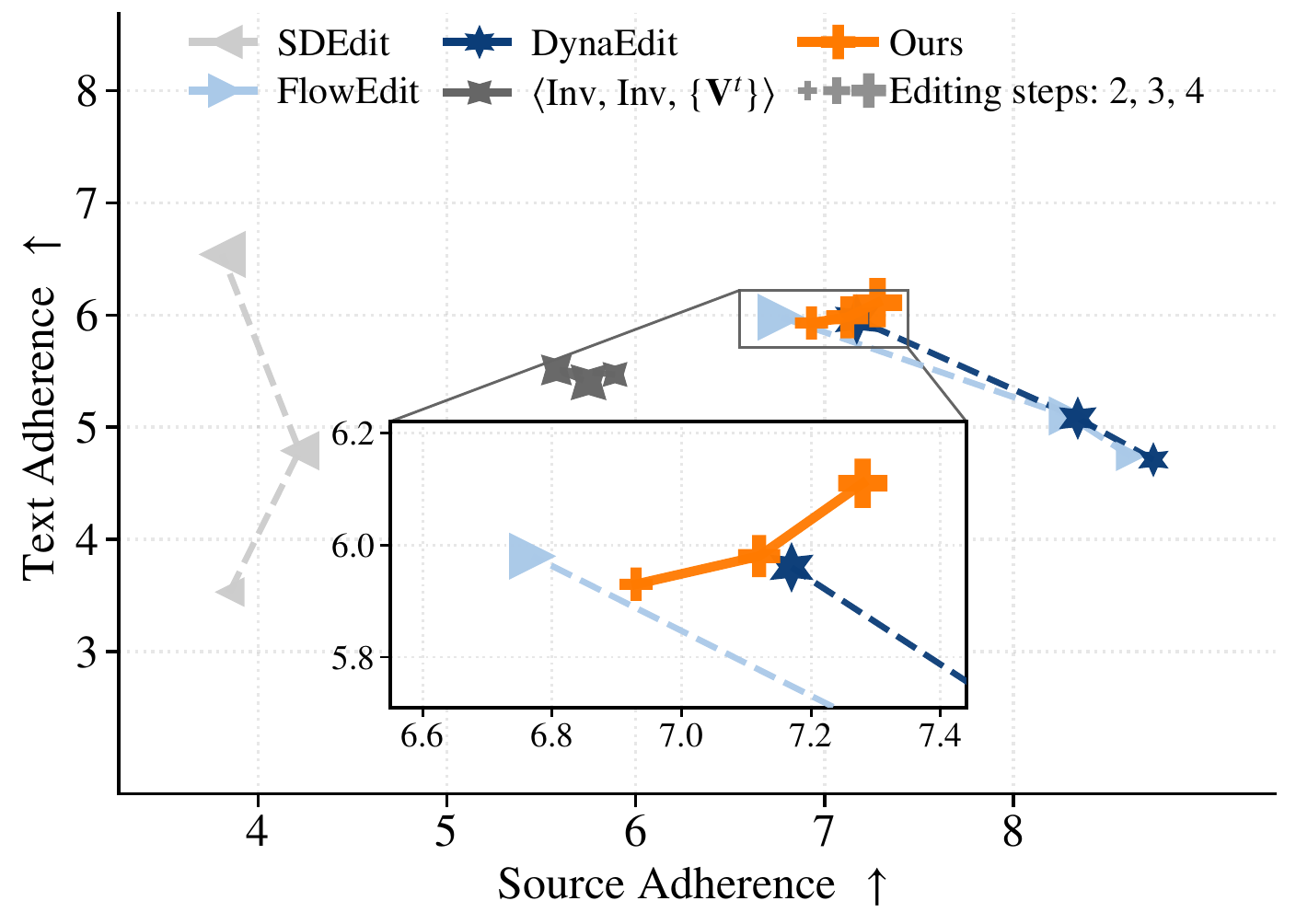}
    \caption{\textbf{Source vs.\ Text adherence across \# of editing steps (indicated by shape size). 
    }
    }
    \label{fig:src_vs_text_editing_steps}
  \end{minipage}\hfill
  \begin{minipage}[t]{0.48\linewidth}
    \vspace{0pt}
    \centering
    \includegraphics[width=\linewidth]{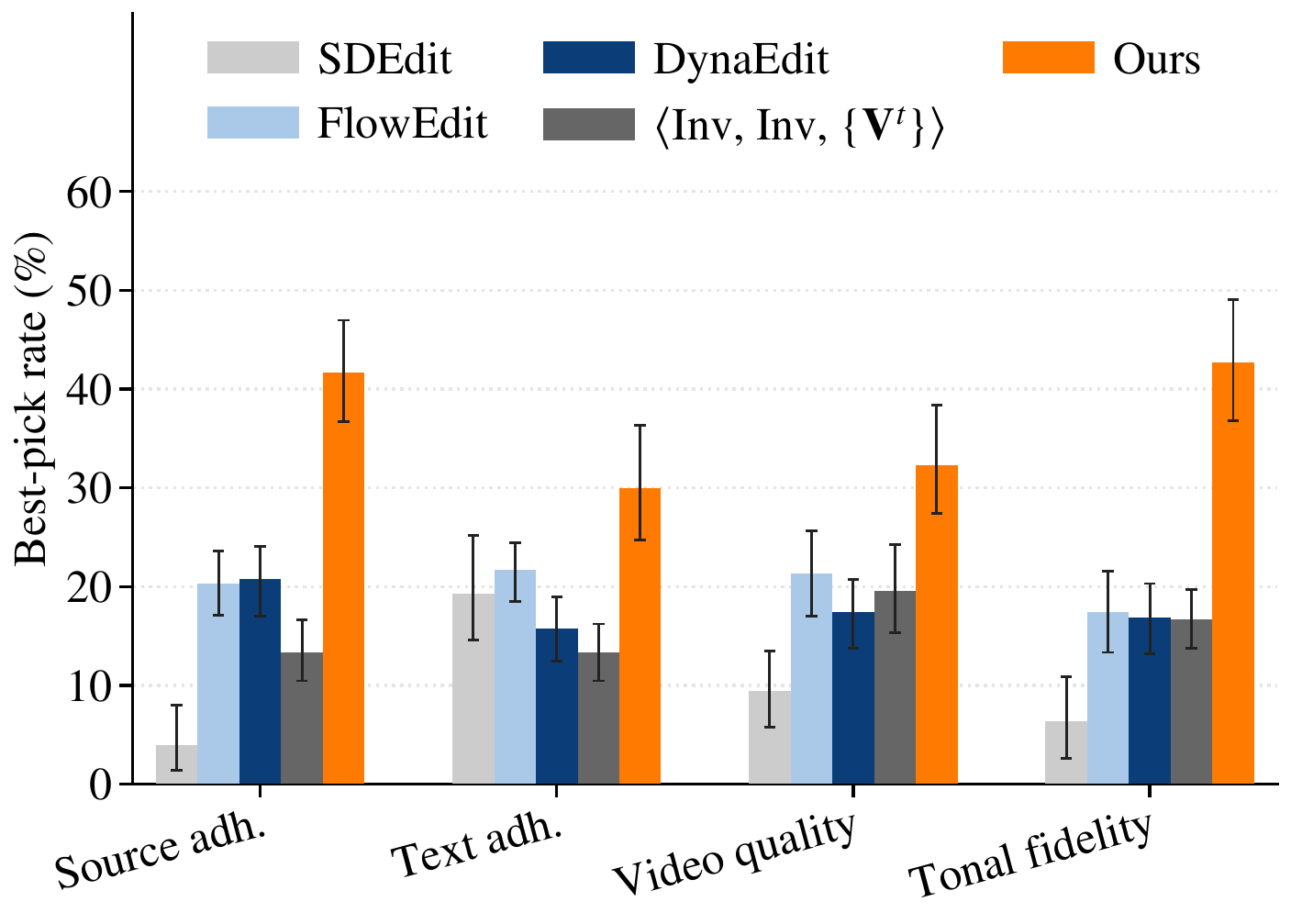}
    \caption{\textbf{User study: best-pick rates per criterion.} }
    \label{fig:user_study}
  \end{minipage}
\end{figure}

\textbf{Default configurations.} All comparisons use a matched number of edit steps per backbone: $4$ on Reward Forcing and Self Forcing, $3$ on CausVid, and $45$ on Wan 2.1, which translate into the per-method NFE counts reported in \Cref{tab:main_comparison}. StreamEdit is also shown at its default 15 steps. SDEdit uses noise level $0.9$. \emph{\recInvInv} uses $S{=}2$ and $b_{\min}{=}22$ ($S=1$ on CausVid); Ours uses $S{=}S_{\text{CtA}}{=}2$, $b_{\min}{=}22$, $b_{\text{CtA}}{=}19$ on Reward Forcing and Self Forcing backbones and $S{=}S_{\text{CtA}}{=}1$ on CausVid. \looseness-1

\begin{figure*}[!t]
    \centering
    \begin{subfigure}[t]{0.48\linewidth}
      \centering
      \includegraphics[width=\linewidth]{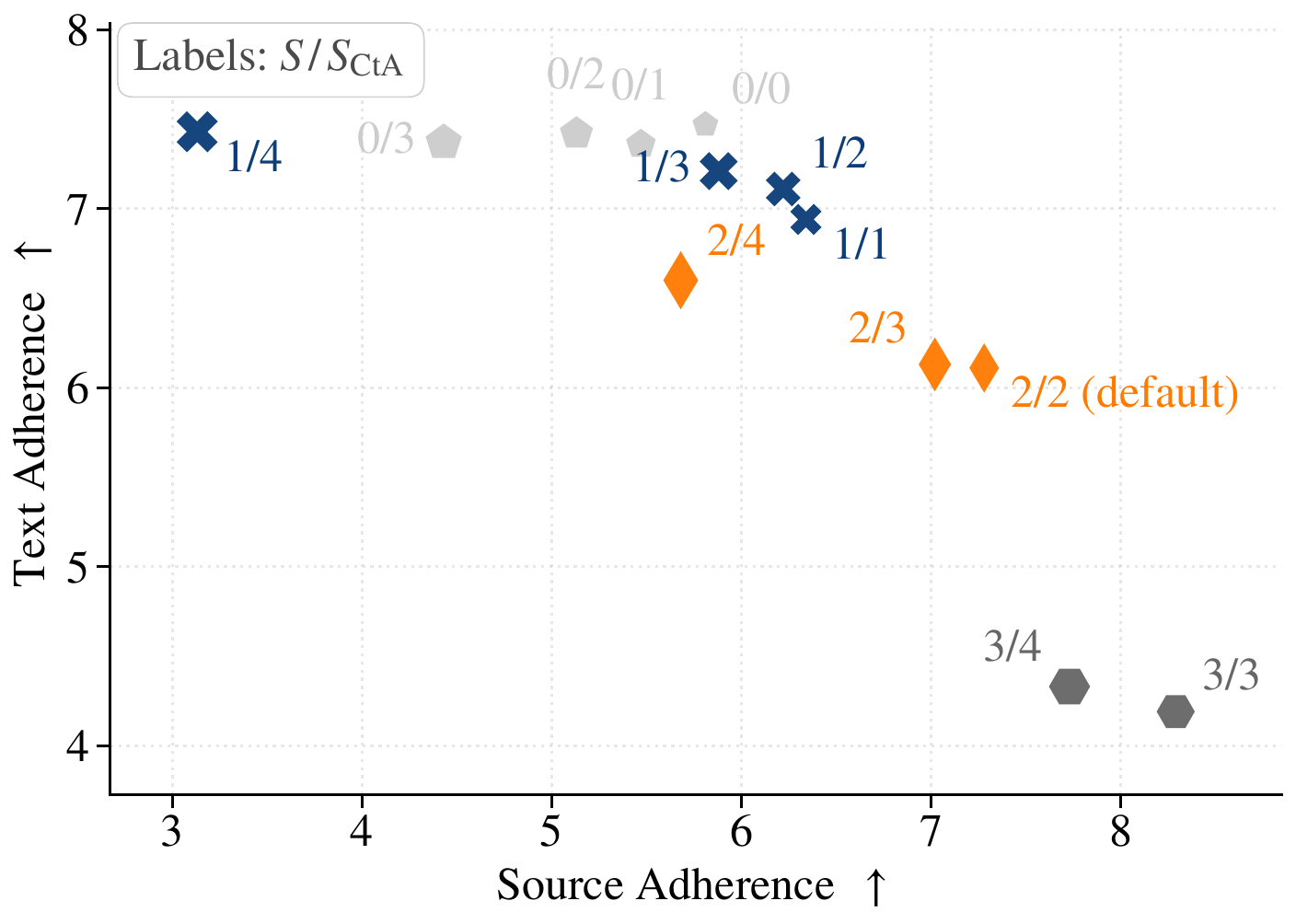}
      \caption{Ablation on $S /S_{\mathrm{CtA}}$. $b_{min}=22, b_{CtA}=19$.}
      \label{fig:abl:S_L:src_text}
    \end{subfigure}\hfill
    \begin{subfigure}[t]{0.48\linewidth}
      \centering
      \includegraphics[width=\linewidth]{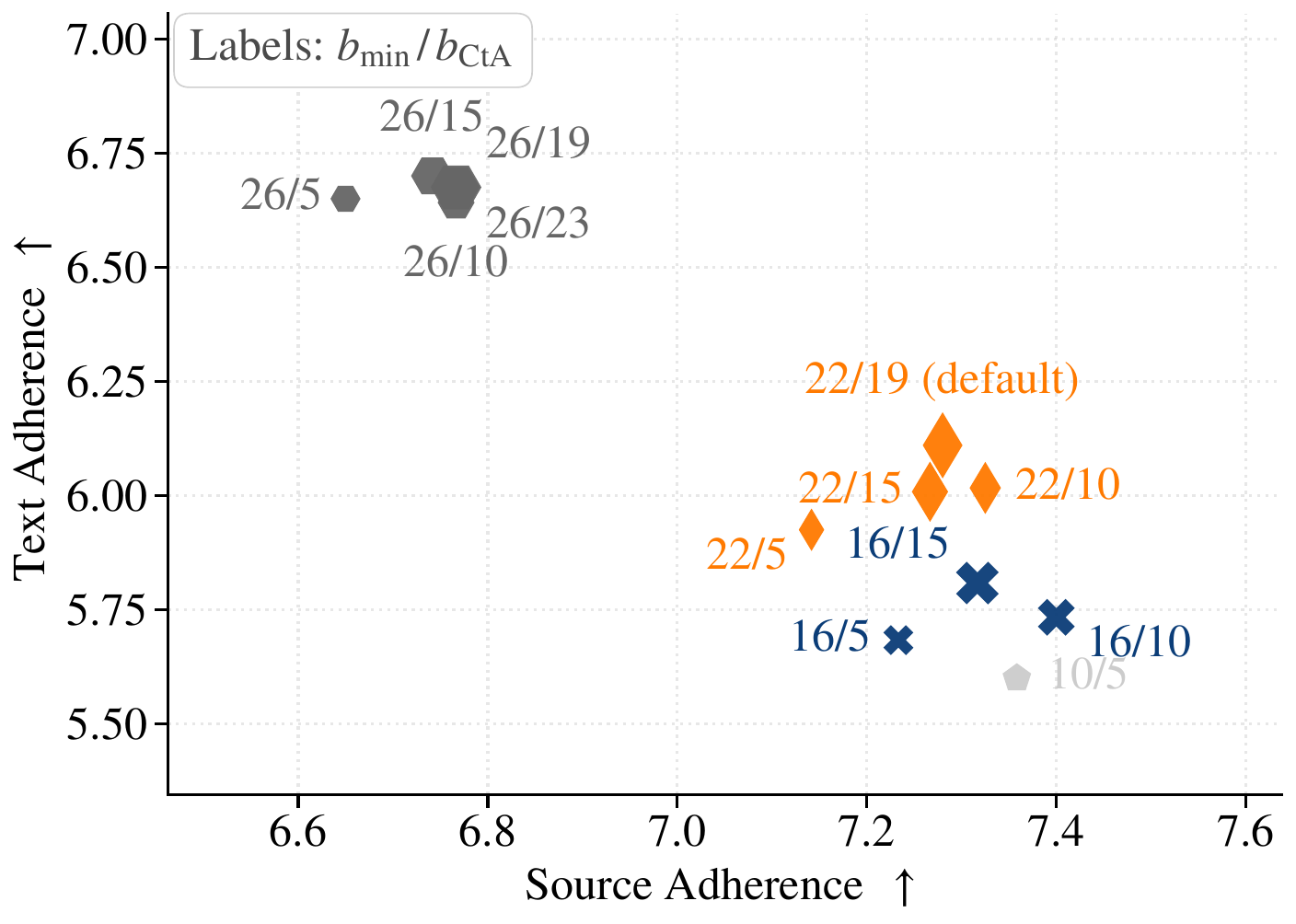}
      \caption{Ablation on $b_{\min}/ b_{\mathrm{CtA}}$. $S=S_{CtA}=2$.}
      \label{fig:abl:mb_cta:src_text}
    \end{subfigure}
    \caption{\textbf{Source vs.\ Text adherence for two ablation grids.} Shape and colour are fixed by $S$ in (a) and by $b_{\min}$ in (b). Our defaults sit at the balanced knee of the source-text trade-off.}
    \label{fig:abl:src_text}
  \end{figure*}

\textbf{Quantitative results.} \Cref{tab:main_comparison} reports results across four backbones. On Reward Forcing, our method leads on source adherence, video quality, TF and TF$_{\text{u}}$. The TF gap over DynaEdit, FlowEdit, and SDEdit confirms that $\mathbf{V}$-injection from a clean forward pass avoids their tonal drift. Against \emph{\recInvInv}, ours improves on every quality axis at fewer NFEs ($6$ vs.\ $9$), confirming inversion is not necessary in this regime. Editing results for the recipes of \Cref{fig:main_figure} are in \Cref{app:appendix_additional_abl}.\looseness=-1

\Cref{tab:main_comparison} also exposes a source--text trade-off: SDEdit reaches the highest text adherence by giving up the source, while FlowEdit and DynaEdit preserve the source at the expense of the edit. On Reward Forcing, our method matches the balance of DynaEdit on both axes at less than half its NFE ($6$ vs.\ $13$) and with markedly higher TF; on CausVid it is the only method that holds both axes (see \Cref{fig:src_vs_text_editing_steps} for source vs.\ text adherence across editing steps). The same trade-off appears in the native bidirectional setting, where FlowEdit and DynaEdit balance both axes thanks to a large step budget, at many times the streaming latency. Under distillation their TF drops; ours holds source adherence, text adherence and TF at the streaming budget, on par with the native methods. 

\textbf{User study.} In a forced-choice study with twenty participants on 33 videos (Reward Forcing backbone; details in~\Cref{app:user_study_protocol}), our method is picked as best on all four criteria (\Cref{fig:user_study}).

\textbf{Qualitative results.} \Cref{fig:visual_comparison_with_sota} shows an object-swapping edit on Reward Forcing. SDEdit loses source identity entirely. FlowEdit and DynaEdit exhibit burning artifacts (red boxes): overexposure and color shift around the subject. \emph{\recInvInv} suppresses burning but introduces spatial inconsistency. Ours preserves source appearance while applying the edit without tonal artifacts. More results, including StreamEdit comparisons on Self Forcing, are in \Cref{app:appendix_additional_eval}.

\textbf{Ablation studies.} We ablate $S$, $S_{\text{CtA}}$, $b_{\min}$ and   $b_{\text{CtA}}$ on Reward Forcing with four editing steps. \Cref{fig:abl:S_L:src_text} varies the injection step $S$ and the commit-then-anchor start $S_{\text{CtA}}$ ($S_{\text{CtA}}{=}4$ disables anchoring). Each step of full $\mathbf{V}$-injection raises source adherence and lowers text adherence, until $S{=}3$ largely suppresses the edit. Anchoring recovers the source at little cost to the text: with $S{=}2$, adding anchoring ($2/2$ against $2/4$) raises source adherence by more than a point while text alignment drops by only half a point, and the gain grows the earlier anchoring starts ($2/3$ lies between the two). Anchoring alone ($S{=}0$) keeps the text but recovers only part of the source, so both phases are needed. \Cref{fig:abl:mb_cta:src_text} varies the block thresholds: injecting into more blocks (smaller $b_{\min}$) trades text for source adherence, but below $b_{\min}{=}22$ the source gain saturates while the text keeps degrading, and $b_{\text{CtA}}$ only nudges text adherence within a group. Our defaults, $S{=}S_{\text{CtA}}{=}2$ and $b_{\min}{=}22$, $b_{\text{CtA}}{=}19$, sit at the knee of both trade-offs. Remaining metrics, visuals and additional ablations are in \Cref{app:appendix_additional_abl}. \looseness=-1

%% file: sections/section2_related_work.tex
\section{Related Work}
\label{sec:related}
 
\textbf{Autoregressive (AR) Video Diffusion Models.} Bidirectional text-to-video models~\cite{wan2025wanopenadvancedlargescale, zheng2024opensorademocratizingefficientvideo, kong2025hunyuanvideosystematicframeworklarge, peng2025opensora20trainingcommerciallevel, ma2025stepvideot2vtechnicalreportpractice} generate all frames jointly but cannot stream. AR video diffusion models \cite{yin2025causvid, huang2025selfforcingbridgingtraintest, liu2025rollingforcingautoregressivelong, lu2025rewardforcingefficientstreaming,  
cui2025selfforcingminutescalehighqualityvideo, kodaira2026streamditrealtimestreamingtexttovideo, yu2025videossmautoregressivelongvideo, hong2025sneakpeekfutureguidedinstructionalstreaming} on the other hand generate videos chunk-by-chunk supporting streaming video generation. 
CausVid~\cite{yin2025causvid} distills a bidirectional Wan \cite{wan2025wanopenadvancedlargescale} teacher into a causal student via asymmetric DMD \cite{yin2024onestepdiffusiondistributionmatching, yin2024improveddistributionmatchingdistillation} with per-chunk noise schedules and KV caching, enabling real-time generation. However, CausVid struggles with exposure bias over long horizon autoregressive video generation.
Self Forcing~\cite{huang2025selfforcingbridgingtraintest} addresses exposure bias by reducing the train--test gap by conditioning on the model's own predictions during training.
A persistent failure across these models is \emph{attention-sink over-dependence}: static initial-frame tokens in the KV cache cause the model to copy early content and suppress motion.
Reward Forcing~\cite{lu2025rewardforcingefficientstreaming} addresses this with EMA sink, continuously updating sink tokens via exponential moving-average fusion of evicted frames, and Re-DMD, which reweights distillation toward high-motion samples. VideoSSM~\cite{yu2025videossmautoregressivelongvideo} replaces the fixed EMA with a state-space model serving as an evolving global memory, dynamically controlling the fusion of local and long-range context. Our work operates \emph{inside} these AR streaming backbones, treating their causal KV cache and few-step predictions as given infrastructure to address the orthogonal question of how to \emph{edit} the stream without training. \looseness=-1

\textbf{Image and Video Editing.} Training-free editing on flow-matching models \cite{lipman2023flowmatchinggenerativemodeling, liu2022flowstraightfastlearning, albergo2023buildingnormalizingflowsstochastic} divides into inversion-based \cite{rout2024semanticimageinversionediting, wang2024taming, deng2024fireflowfastinversionrectified, samuel2025lightningfastimageinversionediting, Avrahami_2025, Shen_2025_ICCV, dalva2024fluxspacedisentangledsemanticediting, ouyang2025proeditinversionbasededitingprompts, cohen2024sliceditzeroshotvideoediting, kara2023raverandomizednoiseshuffling, tokenflow2023, zhang2025adaflowefficientlongvideo, yang2025videograinmodulatingspacetimeattention, fan2024videoshoplocalizedsemanticvideo, wang2025videodirectorprecisevideoediting, bai2024unieditunifiedtuningfreeframework} and inversion-free \cite{kulikov2024flowedit, yoon2025splitflowflowdecompositioninversionfree, kim2025flowaligntrajectoryregularizedinversionfreeflowbased, li2026exacteditingflowbaseddiffusion, jiao2025unieditflowunleashinginversionediting,  lu2026chordeditonesteplowenergytransport, li2025flowdirectortrainingfreeflowsteering, kong2025tamingflowbasedi2vmodels, kulikov2026versatileeditingvideocontent} families.
On the one hand, RF-Solver-Edit~\cite{wang2024taming} and FireFlow~\cite{deng2024fireflowfastinversionrectified} improve inversion via second-order solvers and inject source value features during denoising. Stable Flow~\cite{Avrahami_2025} and ProEdit~\cite{ouyang2025proeditinversionbasededitingprompts} identify critical DiT layers for feature injection.
On the other hand, inversion-free methods bypass inversion.
For example, FlowEdit~\cite{kulikov2024flowedit} and DynaEdit \cite{kulikov2026versatileeditingvideocontent} model editing as a direct ODE driven by the velocity difference between source and target eliminating inversion. All require multiple NFEs for editing, which is incompatible with a 3--4 step distilled student.

For videos, FlowDirector~\cite{li2025flowdirectortrainingfreeflowsteering} introduces Spatially Attentive Flow Correction on top of FlowEdit, which zeros the editing velocity in non-target regions identified by cross-attention masks, and Differential Averaging Guidance, a CFG-inspired \cite{ho2022classifierfreediffusionguidance} signal that steers the edit toward stronger semantic alignment without additional model evaluations.
DynaEdit~\cite{kulikov2026versatileeditingvideocontent} lifts FlowEdit to bidirectional video models and introduces Similarity-Guided Aggregation to fix layout drift and Annealed Noise Correlation to suppress temporal jitter, enabling edits to actions and dynamics. None of these methods address training-free editing on causal, DMD-distilled
AR diffusion models. StreamDiffusionV2~\cite{feng2026streamdiffusionv2streamingdynamicinteractive} integrates SDEdit \cite{meng2022sdeditguidedimagesynthesis} into its real-time pipeline but contributes systems-level scheduling, not editing. Concurrently with our work, StreamEdit~\cite{jiao2026streamedittrainingfreevideoediting} casts editing on few-step streaming video models via source-conditioned noise-to-target generation and dual-branch sampling. \looseness=-1

%% file: sections/section5_conclusion.tex
\section{Conclusion}

We showed that in DMD-distilled autoregressive video diffusion models and their bidirectional teacher, queries and keys remain stable across denoising timesteps while values diverge and encode visual identity. This asymmetry is most pronounced at high noise levels, where few-step autoregressive models operate. A single clean forward pass captures source identity in the values, making inversion unnecessary in this setting. AR-Edit exploits these findings to deliver training-free, inversion-free streaming video editing with strong source preservation and tonal fidelity at minimal latency overhead.

%% file: sections/ack.tex
\begin{ack}
We thank Arijit Ghosh and Lucas Degeorge for valuable discussions, and the anonymous reviewers for their constructive comments. This work was granted access to the HPC resources of IDRIS under the allocations 2025-A0191016617, 2025-AD011016617R1, 2025-AD011016643 and 2026-AD011016643R1 made by GENCI. It was supported by French government funding managed by the National Research Agency (ANR) under the Investments for the Future program (PIA) with the grant ANR-22-EXEN-0002 (eNSEMBLE project). This work benefited from Hi! PARIS and State funding managed by the French National Research Agency (ANR) under the France 2030 program, reference ANR-23-IACL-0005 and the Agence de l'Innovation de Défense (AID) through the Centre Interdisciplinaire d'Études pour la Défense et la Sécurité (CIEDS), under the FakeDetect project (2024). V.Kalogeiton is co-funded by the European Union (ERC, FLASH, 101303557). Views and opinions expressed are however, those of the authors only and do not necessarily reflect those of the European Union or the European Research Council. Neither the European Union nor the granting authority can be held responsible for them.
\end{ack}

%% file: sections/section6_appendix.tex
\appendix
\crefalias{section}{appendix}
\crefalias{subsection}{appendix}

\input{sections/appendix/section_1}
\input{sections/appendix/section_3}
\input{sections/appendix/section_4}
\input{sections/appendix/section_2}
\input{sections/appendix/section_6}

\input{sections/appendix/section_7}
\input{sections/appendix/section_5}

%% file: sections/appendix/section_1.tex
\section{Extended Analysis of the QKV Asymmetry}
\label{app:qkv_extended}

Section~\ref{sec:Attention_Structure} establishes the $\mathbf{QKV}$ asymmetry on the Reward Forcing backbone \cite{lu2025rewardforcingefficientstreaming}. This section expands that analysis along two axes. First, we show that the asymmetry is not specific to a single DMD-distilled autoregressive model by reproducing the analysis on three additional backbones (Section~\ref{app:qkv_models}). Second, we complement the noised-source protocol of Section~\ref{sec:Attention_Structure} by replicating the analysis in two additional settings: during generation from Gaussian noise and during editing  of Algorithm~\ref{alg:v_inject_editing}. The $\mathbf{QKV}$ asymmetry holds in every setting.

\subsection{Generalization Across Model Families}
\label{app:qkv_models}

We analyze three backbones beyond Reward Forcing, examining them in order of increasing distance from the DMD-distilled autoregressive setting discussed in the main paper: first CausVid~\cite{yin2025causvid}, a second DMD-distilled autoregressive student, then its DMD-distilled bidirectional version, and finally the non-distilled bidirectional Wan~2.1~1.3B teacher~\cite{wan2025wanopenadvancedlargescale}. Together, the four models cover all combinations of causal versus bidirectional attention and DMD-distilled versus non-distilled, and each step of the analysis removes one ingredient of our setting, first causality and then distillation. We follow the same analysis protocol as Section~\ref{sec:Attention_Structure}.

\begin{figure*}[t]
    \centering
    \includegraphics[width=1.0\linewidth]{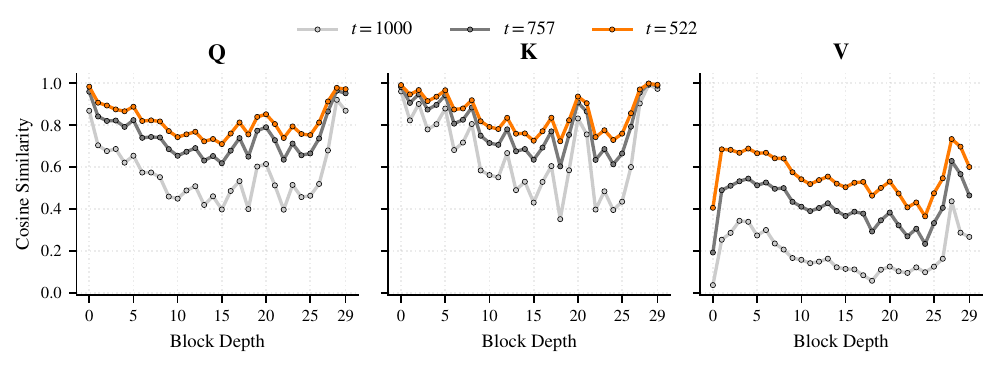}%
    \caption{\textbf{$\mathbf{QKV}$ asymmetry on CausVid (DMD-distilled, autoregressive).} Cosine similarity of $\mathbf{Q}$, $\mathbf{K}$, and $\mathbf{V}$ at each timestep \(t \in \{1000, 757, 522\}\) to their \(t{=}0\) (clean source) counterparts, averaged across 120 videos.}
\label{fig:qkv_causvid_ar}
\end{figure*}

\begin{figure*}[t]
    \centering
    \includegraphics[width=1.0\linewidth]{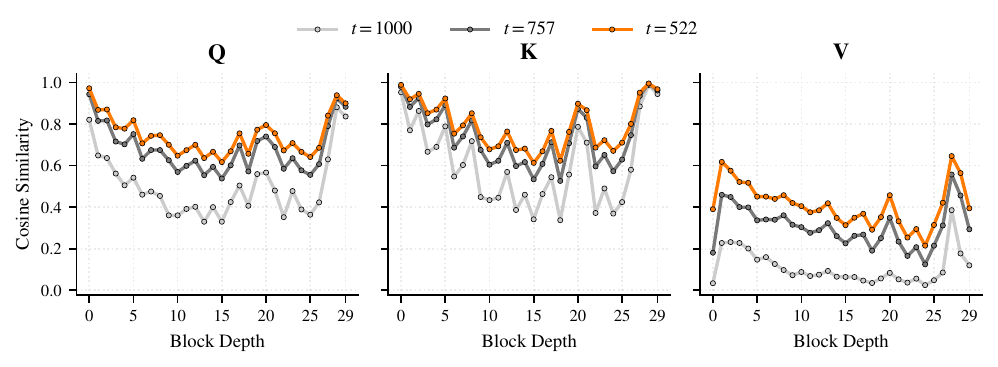}%
    \caption{\textbf{$\mathbf{QKV}$ asymmetry on CausVid-Bidirectional (DMD-distilled, bidirectional).} Cosine similarity of $\mathbf{Q}$, $\mathbf{K}$, and $\mathbf{V}$ at each timestep \(t \in \{1000, 757, 522\}\) to their \(t{=}0\) (clean source) counterparts, averaged across 120 videos. $\mathbf{Q}$ and $\mathbf{K}$ retain high similarity to the clean anchor while $\mathbf{V}$ diverges, matching its autoregressive version.}
\label{fig:qkv_causvid_bidir}
\end{figure*}

\textbf{DMD-distilled students beyond Reward Forcing.} Figure~\ref{fig:qkv_causvid_ar} repeats the analysis on CausVid, a DMD-distilled autoregressive student of the Wan~2.1 teacher, measured at its three training timesteps ($t \in \{1000, 757, 522\}$). CausVid reproduces the $\mathbf{QKV}$ asymmetry, showing that the observation is not unique to one distilled model. Figure~\ref{fig:qkv_causvid_bidir} then removes causality: the bidirectional DMD-distilled model released by the CausVid authors matches the behavior of its autoregressive counterpart.

\begin{figure*}[t]
    \centering
    \includegraphics[width=1.0\linewidth]{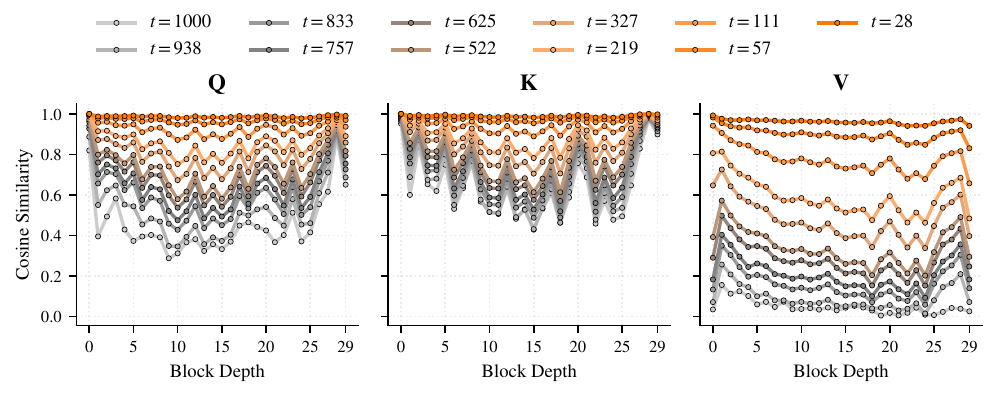}%
    \caption{\textbf{$\mathbf{QKV}$ asymmetry on the non-distilled bidirectional Wan~2.1~1.3B teacher.} Cosine similarity of $\mathbf{Q}$, $\mathbf{K}$, and $\mathbf{V}$ to their \(t{=}0\) (clean source) counterparts, averaged over 120 videos.}
\label{fig:qkv_teacher}
\end{figure*}

\textbf{Non-distilled bidirectional teacher.} Figure~\ref{fig:qkv_teacher} removes distillation, by reporting the analysis on the Wan~2.1~1.3B teacher, a bidirectional multi-step video diffusion model, measured at the timesteps of the Reward Forcing schedule ($t \in \{1000, 938, 833, 625\}$), the CausVid schedule ($t \in \{1000, 757, 522\}$), and additional mid- and low-noise timesteps ($t \in \{327, 219, 111, 57, 28\}$) covering the full noise range. The teacher exhibits the same asymmetry as the students: at $t{=}1000$, block~0, $\mathbf{Q}$  and $\mathbf{K}$  reach similarities of $0.82$ and $0.95$ to the clean anchor while $\mathbf{V}$ sits at $0.04$, and the first two match the values of the bidirectional student exactly, showing that the students inherit the asymmetry directly from the teacher. Beyond confirming the $\mathbf{QKV}$ asymmetry on the non-distilled bidirectional model, the full noise range reveals that its strength depends on the noise level: toward low noise the gap between values and the clean anchor narrows and has largely vanished by $t{=}28$, where all three features approach their clean counterparts ($t{=}0$).

Both students train on and operate at these high-noise timesteps: all of Reward Forcing's training timesteps satisfy $t \geq 625$ and CausVid's satisfy $t \geq 522$, so their entire sampling schedules lie in the regime where the inherited asymmetry is at its strongest. The distilled schedules never visit the low-noise region where the asymmetry weakens, so the values cached from a single clean pass at $t{=}0$ are a valid source anchor at every step, while inversion, on which multi-step bidirectional video diffusion models can rely, is ill-defined for few-step models (Section~\ref{sec:Preliminaries}). The distilled autoregressive regime is thus where the $\mathbf{QKV}$ asymmetry is both most pronounced and most needed.

\subsection{Robustness Across Inference Regimes}
\label{app:qkv_inference}

Extending the noised-source analysis of Section~\ref{sec:Attention_Structure}, we replicate the $\mathbf{QKV}$ asymmetry analysis on the Reward Forcing backbone, as in the main paper, in two additional settings: during generation (denoising) from Gaussian noise, and during editing  (Algorithm~\ref{alg:v_inject_editing}). The asymmetry holds in both.

\begin{figure*}[t]
    \centering
    \includegraphics[width=1.0\linewidth]{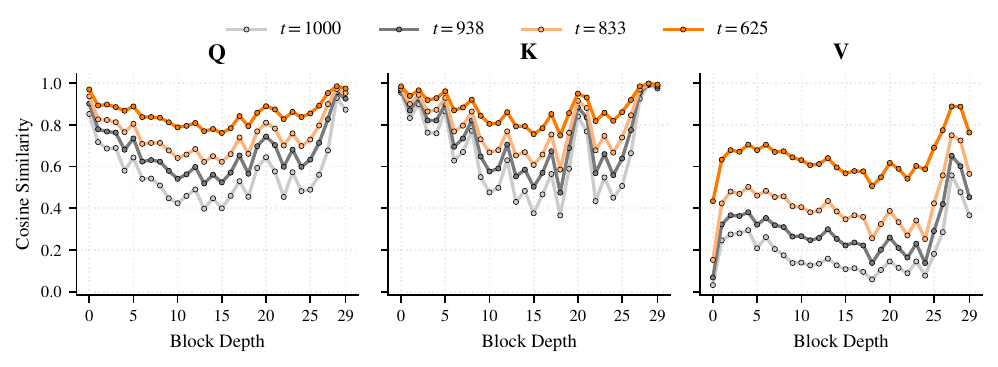}%
    \caption{\textbf{$\mathbf{QKV}$ asymmetry during generation from noise on Reward Forcing.} Cosine similarity of $\mathbf{Q}$, $\mathbf{K}$, and $\mathbf{V}$ at each timestep \(t \in \{1000, 938, 833, 625\}\) to their \(t{=}0\) (clean source) counterparts when generating from pure Gaussian noise with the backbone's native predict-then-renoise sampler, averaged across 120 videos.}
\label{fig:qkv_generation}
\end{figure*}

\textbf{Generation-time analysis.} Starting from Gaussian noise, we run a full denoising on the benchmark prompts with the backbone's native predict-then-renoise sampler and extract at each timestep the post-RoPE queries and keys and the pre-attention values, as in the main analysis. Figure~\ref{fig:qkv_generation} shows that the $\mathbf{QKV}$ asymmetry also holds in this inference regime.

\begin{figure*}[t]
    \centering
    \includegraphics[width=1.0\linewidth]{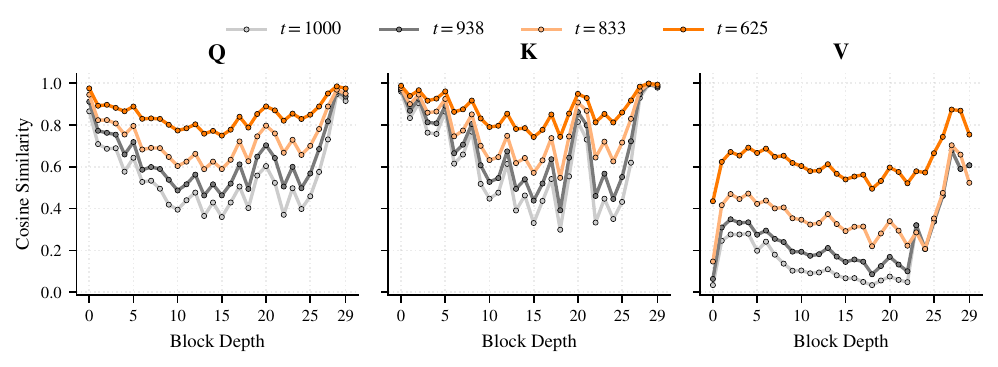}%
    \caption{\textbf{$\mathbf{QKV}$ asymmetry during editing on Reward Forcing.} Cosine similarity of $\mathbf{Q}$, $\mathbf{K}$, and $\mathbf{V}$ at each timestep \(t \in \{1000, 938, 833, 625\}\) to their \(t{=}0\) (clean source) counterparts during the editing of Algorithm~\ref{alg:v_inject_editing}, averaged across 120 videos.}
\label{fig:qkv_editing}
\end{figure*}

\textbf{Editing-time analysis.} We repeat the $\mathbf{QKV}$ analysis during the editing of Algorithm~\ref{alg:v_inject_editing} on the benchmark with the default parameters of Appendix~\ref{app:appendix_imp_details}, where the source values cached at $t{=}0$ are injected during denoising. Figure~\ref{fig:qkv_editing} shows that the asymmetry persists in this regime: queries and keys remain close to the $t{=}0$ anchor while values diverge, matching the noised-source analysis of Figure~\ref{fig:fig_cs_across_timesteps}.

%% file: sections/appendix/section_3.tex
\section{Implementation Details}
\label{app:appendix_imp_details}

All methods run at $480\times832$ resolution. We provide evaluations on four backbones: the non-distilled bidirectional Wan~2.1 1.3B~\cite{wan2025wanopenadvancedlargescale}, and three DMD-distilled autoregressive students of it, Reward Forcing~\cite{lu2025rewardforcingefficientstreaming}, CausVid~\cite{yin2025causvid} and Self Forcing ~\cite{huang2025selfforcingbridgingtraintest}, which run with causal attention and chunked KV caching. \emph{\recInvInv} is run on Wan 2.1, Reward Forcing and CausVid, Ours on all three AR backbones. Next, we detail the configurations used for our method and all compared approaches, on each backbone in turn.

\textbf{SDEdit.}
Following~\cite{meng2022sdeditguidedimagesynthesis, feng2026streamdiffusionv2streamingdynamicinteractive}, we forward-diffuse the source
video to a certain noise level under the backbone's schedule, then denoise under the target prompt: over the whole input on the bidirectional backbone, and chunk-by-chunk with the standard streaming loop and shared KV cache on the two AR backbones. For \Cref{tab:main_comparison} we set the noise level to $0.9$ on all three backbones, and vary it on Reward Forcing in \Cref{fig:src_vs_text_editing_steps} resulting in different editing steps.

\textbf{FlowEdit and DynaEdit.}
FlowEdit~\cite{kulikov2024flowedit} and DynaEdit~\cite{kulikov2026versatileeditingvideocontent} integrate the velocity-field
difference between source- and target-prompted trajectories along a
direct ODE, bypassing noise injection entirely. We do not use the SDEdit style refinement stage of the methods in our implementation. At each step we evaluate both velocity fields and integrate
their difference, transporting the source toward the target without
passing through pure noise. On the bidirectional backbone we run both methods with the maximum number of editing steps set to $45$. On the AR backbones we run the ODE chunk-by-chunk with the same KV-cache discipline as standard generation, with maximum number of editing steps set to $4$ on Reward Forcing and $3$ on CausVid. $4$ editing steps on Reward Forcing at two network passes per step plus one pass for the context rerun, yields $9$ NFEs in \Cref{tab:main_comparison} in the case of FlowEdit. On every backbone, DynaEdit aggregates $3$ samples in its similarity-guided aggregation at the first step and a single sample at every later step, thus resulting in 13 NFEs on Reward Forcing in \Cref{tab:main_comparison}.

\textbf{StreamEdit.}
We run StreamEdit~\cite{jiao2026streamedittrainingfreevideoediting} with $4$ editing steps, matching the budget of our method, and keep all other hyperparameters at the values of the official paper. StreamEdit uses $4$ editing steps at two network passes per step, plus two cache-update passes, yielding $10$ NFEs in \Cref{tab:main_comparison}. Additionally, we report StreamEdit at its default 15 edit steps resulting in 32 NFEs.

\textbf{\emph{\recInvInv}.}
We perform inversion on the source latents to
obtain noisy references, then start denoising initialized from
those latents. For \Cref{tab:main_comparison} we set the number of editing steps to $4$, $4$ inversion steps, $S{=}2$, minimum injection block to $22$ and one context rerun resulting in $9$ NFEs on Reward Forcing. On CausVid we set $S{=}1$ with $3$ editing steps and the same minimum injection block. On the bidirectional Wan~2.1 1.3B model, we use 45 inversion and 45 editing steps (90 NFE in \Cref{tab:main_comparison}), with $S{=}2$ and $b_{\min}{=}24$.

\input{sections/algorithm1}

\textbf{Ours.} \Cref{alg:v_inject_editing} gives the full pseudo-code of our method.
Our method uses the same $\mathbf{V}$-injection mechanism but starts
from random Gaussian noise instead of inverted noise. $\mathbf{V_{src}}$ is computed once from a single clean forward pass at
$t{=}0$. Thus, ours uses one pass for $\mathbf{V_{src}}$ extraction, $4$ for denoising, and one for the context rerun, total of $6$ NFEs in \Cref{tab:main_comparison}.

As described in \Cref{sec:method}, the editing schedule has two thresholds: 
  $\mathbf{V_{src}}$ replaces the target's at every token of the chunk during steps
  $k < S$, and is applied only outside the edit region, through the soft mask
  $M$, during steps $k \geq S_{\text{CtA}}$ with $S_{\text{CtA}} \geq S$. Steps
  that fall between the two, if any, are left without $\mathbf{V_{src}}$ injection. The
  mask is derived from the cross-attention of both the source and the target
  pass, and the target side is only informative once the layout of the edit has
  formed: at the first steps the target sample is still close to noise, and a
  mask read from its cross-attention marks no coherent region. We thus
  inject $\mathbf{V_{src}}$ without a mask until the layout is generated, and then
  switch to masked injection; by default $S_{\text{CtA}} = S = 2$ with
  $K = 4$ on Reward Forcing and Self Forcing, and $S_{\text{CtA}} = S = 1$ with $K = 3$ on CausVid; the block thresholds and the text boost described next are shared by all AR backbones.  We ablate our method with and without CtA in \Cref{app:appendix_additional_abl}.

We obtain $M$ from the model's cross-attention, which localises where each prompt word acts, following~\cite{cai2025ditctrlexploringattentioncontrol, ouyang2025proeditinversionbasededitingprompts}. We align the source and target prompts with the Needleman--Wunsch algorithm~\cite{needleman1970general} to identify the words that differ between them, and read off where these words attend: on the source side from the clean source pass, on the target side from the current denoising pass. A token belongs to the edit region if the differing words attend to it above a threshold. Raw attention maps are noisy, so we keep only the largest connected regions per latent frame, dilate them, and feather the boundary. By default, the mask is accumulated over the cross-attention of blocks 0--19 ($b_{CtA}=19$), while $\mathbf{V_{src}}$ is injected in blocks 23--29 ($b_{min}=22$), so the mask is complete before it is needed within the same pass. For the first chunk, the final anchor step injects the masked $\mathbf{V_{src}}$ in blocks 15--29 ($b_{\min}=14$), reusing the mask built at the preceding step. We ablate $b_{min}$ and $b_{CtA}$ in \Cref{app:appendix_additional_abl}.

Moreover, the target-side words identified by the prompt alignment above, i.e.\ the words that carry the edit, receive a text boost: in every cross-attention layer, their unnormalized attention weights are multiplied by a constant factor $s$ and the attention row is renormalized ~\cite{kim2025reflextextguidededitingreal}. We use $s=4$ by default (see \Cref{app:appendix_additional_abl} for an ablation). The text boost is applied at all editing steps.

%% file: sections/algorithm1.tex
\begin{algorithm}[t]
\caption{AR-Edit: Autoregressive Video Editing without Inversion.}
\label{alg:v_inject_editing}
\small
\begin{algorithmic}[1]
\Require Incoming source chunks $\{\mathbf{x}_{\text{src}}^{(n)}\}_{n=0}^{N-1}$ of $F$ frames; source / target prompts $c_{\text{src}}, c_{\text{tar}}$; distilled model $G_\theta$; injection budget $S\!\in\!\{0,\dots,K\}$; min block id $b_{\min}\!\in\!\{-1,\dots,L-1\}$; anchor start $S_{\text{CtA}}\!\geq\!S$; mask block $b_{\text{CtA}}\!<\!b_{\min}$; text-boost scale $s$.
\Ensure Edited chunks $\{\hat{\mathbf{x}}^{(n)}\}_{n=0}^{N-1}$.

\State $\mathcal{K}_{\text{gen}},\,\mathcal{X}_{\text{gen}} \gets \emptyset$ \Comment{Generation-side self / cross-attn KV caches; \emph{persist} across chunks for AR context.}
\State $\mathcal{T}_{\text{rem}} \gets$ tokens of $c_{\text{src}}$ not in $c_{\text{tar}}$, $\mathcal{T}_{\text{new}} \gets$ tokens of $c_{\text{tar}}$ not in $c_{\text{src}}$ \Comment{See \Cref{app:appendix_imp_details}.}

\For{$n=0,1,\dots,N-1$}

  \LComment{\textbf{Step 1.} Per-chunk $\mathbf{V}$-extraction from the clean source (single forward at $t{=}0$).}
  \State $\mathcal{K}_{\text{src}} \gets \emptyset$ \Comment{Source-side cache reset per chunk so each chunk's $\mathbf{V}$ is harvested in isolation.}
  \State $G_\theta\!\bigl(\mathbf{x}_{\text{src}}^{(n)},\,t{=}0,\,c_{\text{src}};\,\mathcal{K}_{\text{src}}\bigr)$, where inside the forward:
  \State \hskip\algorithmicindent at blocks $l\leq b_{\text{CtA}}$: $A_{\text{rem}}^{(n)} \gets$ cross-attn of the $\mathcal{T}_{\text{rem}}$ columns
  \State \hskip\algorithmicindent at blocks $l>b_{\min}$: $\mathbf{V}_{\text{src}}^{(n,l)} \gets$ value tensor \Comment{capture $\mathbf{V}$}

  \LComment{\textbf{Step 2.} Denoise the chunk from random noise with $\mathbf{V}$-injection via Commit-then-Anchor (CtA).}
  \State Sample $\mathbf{z}\sim\mathcal{N}(\mathbf{0},\mathbf{I})$
  \For{$k=0,1,\dots,K-1$}
    \State $\textit{inject} \gets (k < S)$;\quad $\textit{anchor} \gets (k \geq S_{\text{CtA}})$
    \State $\hat{\mathbf{x}}_0 \gets G_\theta\!\bigl(\mathbf{z},\,t_k,\,c_{\text{tar}};\,\mathcal{K}_{\text{gen}},\mathcal{X}_{\text{gen}}\bigr)$, where inside the forward, in block order:
    \State \hskip\algorithmicindent every cross-attn layer: add $\log s$ to the logits of the $\mathcal{T}_{\text{new}}$ columns \Comment{text boost}
    \State \hskip\algorithmicindent $\mathbf{K}^{(n,l)}, \mathbf{V}^{(n,l)}$ written into $\mathcal{K}_{\text{gen}}$ \Comment{past-chunk attention sees \emph{native} $\mathbf{K}$,$\mathbf{V}$}
    
    \State \hskip\algorithmicindent at blocks $l\leq b_{\text{CtA}}$, if $\textit{anchor}$: $A_{\text{new}}^{(n)} \gets$ cross-attn of the $\mathcal{T}_{\text{new}}$ columns
    \State \hskip\algorithmicindent at block $b_{\text{CtA}}$, if $\textit{anchor}$: $M \gets \textsc{BuildMask}\bigl(A_{\text{rem}}^{(n)}, A_{\text{new}}^{(n)}\bigr)$ \Comment{mask for CtA, \Cref{app:appendix_imp_details}}
    \State \hskip\algorithmicindent \textbf{at blocks $l>b_{\min}$, if $\textit{inject}$:} $\mathbf{V}^{(n,l)} \gets \,\mathbf{V}_{\text{src}}^{(n,l)}$ \Comment{replace V in attention only; commit}
    \State \hskip\algorithmicindent \textbf{at blocks $l>b_{\min}$, if $\textit{anchor}$:} $\mathbf{V}^{(n,l)} \gets M \odot \mathbf{V}^{(n,l)} + (1-M)\odot \mathbf{V}_{\text{src}}^{(n,l)}$ \Comment{anchor}
    \If{$k<K-1$}
      \State $\sigma_{k+1}\gets\sigma(t_{k+1})$;\quad sample $\boldsymbol{\epsilon}\sim\mathcal{N}(\mathbf{0},\mathbf{I})$
      \State $\mathbf{z} \gets (1-\sigma_{k+1})\,\hat{\mathbf{x}}_0 + \sigma_{k+1}\,\boldsymbol{\epsilon}$ \Comment{ predict-then-renoise update}
    \EndIf
  \EndFor
  \State $\hat{\mathbf{x}}^{(n)} \gets \hat{\mathbf{x}}_0$

  \LComment{\textbf{Step 3.} Clean-context rerun: overwrite the chunk's slot in $\mathcal{K}_{\text{gen}}$ with clean $\mathbf{K}$,$\mathbf{V}$.}
  \State $G_\theta\!\bigl(\hat{\mathbf{x}}^{(n)},\,t{=}t_{\text{ctx}},\,c_{\text{tar}};\,\mathcal{K}_{\text{gen}},\mathcal{X}_{\text{gen}}\bigr)$ \texttt{[ updates\;}$\mathcal{K}_{\text{gen}}$\texttt{]} 
  \State \textbf{yield} $\hat{\mathbf{x}}^{(n)}$
\EndFor
\end{algorithmic}
\end{algorithm}

%% file: sections/appendix/section_4.tex
\section{Benchmark Construction Details}
\label{app:benchmark}

Our benchmark consists of $120$ hand-curated video clips sourced from Pexels~\cite{pexels} under the Pexels License (free use, no attribution required), distributed over $10$ editing categories with $12$ clips each. Every clip is manually selected against a set of editorial constraints, and every prompt pair is verified and refined by hand.

\textbf{Category design.} The benchmark covers four editing categories following DynaEdit \cite{kulikov2026versatileeditingvideocontent}, object insertion, object swapping, motion modification, and global scene transformation, and six attribute categories following PnPInversion~\cite{ju2023direct}: change of color, material, pose, content, background, and style. We construct the benchmark category first: for each category we define the intended edit (e.g., change how the subject moves; transform the whole scene by season, time of day, or weather; replace the main subject) and, from it, derive the search queries describing footage on which that edit is performable (e.g., for motion modification, a subject in clear motion, such as flowing water, a running animal, or a waving flag). Clips are then collected to match, so every clip supports the edit its category tests.

\textbf{Selection rules.} Every candidate clip must satisfy a set of editorial constraints that keep the benchmark focused on generic, subject-centric editing: no people as the primary subject, avoiding identity-preservation concerns; no drone or aerial footage, which is dominated by global parallax; a single centered subject (at most two), so the intended edit region is unambiguous; visible subject motion, ruling out static imagery panned by the camera; and medium or wide framing, so both the edit region and its context remain visible.

\textbf{Prompt construction.} Every clip is paired with (i)~a source caption describing the visible content and (ii)~a target prompt that modifies only the lexical span corresponding to the intended edit. Source captions are first drafted automatically by Qwen2.5-VL-7B~\cite{bai2025qwen25vltechnicalreport} and are then human-refined against the video to their final caption. Target prompts are hand-authored per clip on top of the refined source caption, editing only the span of the intended change (e.g., ``A \textit{tiger} walks confidently through its enclosure, exploring and surveying its surroundings under daylight.'' → ``A \textit{lion} walks confidently through its enclosure, exploring and surveying its surroundings under daylight.'').

\textbf{Diversity and statistics.} The $120$ clips come from $88$ unique Pexels uploaders, which, combined with the per-subject and animal caps, prevents any single producer or subject class from dominating. Table~\ref{tab:pexels_bench_dist} reports per-category statistics: aesthetic score (LAION Aesthetic V2 \cite{schuhmann2022iap}), mean optical-flow magnitude \cite{1763974.1764031}, flow residual after subtracting global camera motion, fraction of moving pixels, and source clip duration, averaged over the $12$ clips of each category.

\begin{table}[t]
  \centering
  \small
  \caption{\textbf{Per-category distribution of the Pexels benchmark.} All metrics are means over the $12$ clips per category. \emph{Aesthetic} is the LAION Aesthetic V2 score~\cite{schuhmann2022iap}; \emph{Motion} is the mean dense optical-flow magnitude~\cite{1763974.1764031} between frames at $10\%$ and $60\%$ of the clip duration; \emph{Flow res.} subtracts the mean flow vector, isolating object from camera motion; \emph{\% mov.} is the fraction of pixels with flow residual above $1.5$; \emph{Dur.} is the source clip duration in seconds.}
  \label{tab:pexels_bench_dist}
  \vspace{4pt}
  \setlength{\tabcolsep}{4pt}
  \begin{tabular}{@{}lcccccc@{}}
    \toprule
    Category & Videos & Aesthetic & Motion & Flow res. & \% mov. & Dur. (s) \\
    \midrule
    Motion modification           & 12 & 5.10 &  4.33 &  4.88 & 61.4 & 20.34 \\
    Global scene transformation   & 12 & 5.39 &  8.74 &  8.64 & 76.6 & 17.82 \\
    Object insertion              & 12 & 5.19 & 10.21 &  7.77 & 86.9 & 34.63 \\
    Object swapping               & 12 & 4.94 &  7.34 &  7.33 & 70.8 & 10.53 \\
    Change attribute: pose        & 12 & 5.11 &  8.48 &  9.41 & 88.5 & 18.15 \\
    Change attribute: material    & 12 & 5.05 &  9.53 &  8.86 & 77.2 & 20.68 \\
    Change style                  & 12 & 5.06 &  8.83 &  9.12 & 84.6 & 25.36 \\
    Change background             & 12 & 5.24 &  7.30 &  7.25 & 71.2 & 15.13 \\
    Change attribute: content     & 12 & 5.11 &  8.76 &  9.34 & 78.4 & 15.08 \\
    Change attribute: color       & 12 & 5.12 &  9.29 &  7.68 & 60.6 & 31.13 \\
    \midrule
    \textbf{Total}                & \textbf{120} & \textbf{5.13} & \textbf{8.28} & \textbf{8.03} & \textbf{75.6} & \textbf{20.89} \\
    \bottomrule
  \end{tabular}
\end{table}

%% file: sections/appendix/section_2.tex
\section{Additional Evaluation Results}
\label{app:appendix_additional_eval}

\begin{figure*}[!t]
  \centering
  \includegraphics[width=1.0\linewidth]{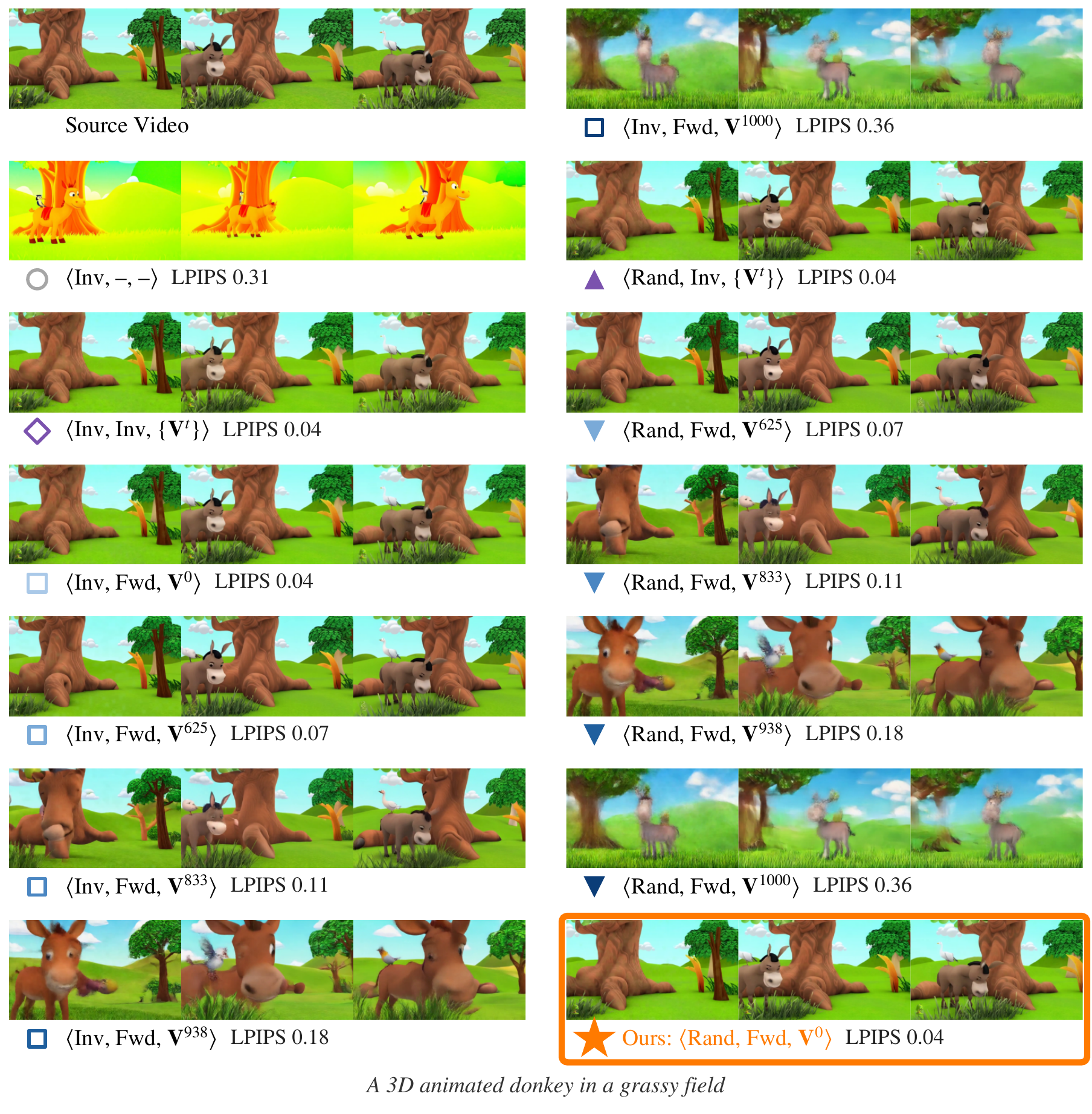}
\caption{\textbf{Qualitative reconstructions for the recipes of \Cref{fig:main_figure}.} Top row: input frames. Each following row is one recipe, with the marker of \Cref{fig:main_figure}, run at the smallest pre-denoising NFE its definition allows; the fixed-$t$ recipes appear once per extraction timestep $t$. Reading top to bottom, the source appearance is lost by inversion alone, restored by every $\mathbf{V}$-injection variant, and fades as the extraction timestep $t$ moves away from the clean source. Ours recovers the input faithfully.}
  \label{fig:recon:grid}
\end{figure*}

\textbf{Visual results of reconstruction.}
\Cref{fig:recon:grid} shows reconstructions for the recipes of \Cref{fig:main_figure}, each run at the smallest pre-denoising NFE its recipe allows. The pattern matches the quantitative study: pure inversion diverges from the source, whereas every $\mathbf{V}$-injection variant reconstructs it faithfully as long as $\mathbf{V}$ comes from the clean source or the inversion path. Once $\mathbf{V}$ comes from a forward pass, the initial noise does not matter: the rows with inverted and random initialization are visually identical. For the fixed-$t$ recipes, fidelity degrades as $t$ grows. Ours, which starts from random noise and takes $\mathbf{V}$ from the clean source at $t{=}0$, matches or exceeds the inversion-based recipes with a single pre-denoising NFE and is indistinguishable from the input.

\textbf{Per-category results.} \Cref{tab:per_category} breaks the Reward Forcing results of \Cref{tab:main_comparison} down by edit category. Our method is best or second-best in source adherence in nine of the ten categories (all but color, where DynaEdit and FlowEdit lead) and has the highest text adherence on motion, material, style and content edits; its weakest category is global scene transformation, where every source-preserving method drops, in line with the trade-off discussed in \Cref{sec:limitations}.

\begin{table*}[t]
  \centering
  \scriptsize
  \caption{\textbf{Per-category source and text adherence on our benchmark} (Reward Forcing backbone, default configurations of \Cref{tab:main_comparison}). Each cell reports Source/Text adherence, mean over the 12 clips of the category. Best per column and metric in \textbf{bold}, second-best \underline{underlined}.}
  \label{tab:per_category}
  \setlength{\tabcolsep}{2.4pt}
  \renewcommand{\arraystretch}{1.12}
  \begin{tabular}{@{}l cccc cccccc c@{}}
    \toprule
    & \multicolumn{4}{c}{\textbf{Dynamics edits}} & \multicolumn{6}{c}{\textbf{Attribute edits}} & \\
    \cmidrule(lr){2-5} \cmidrule(lr){6-11}
    Method & Motion & Global & Insertion & Swap & Pose & Material & Style & Background & Content & Color & All \\
    \midrule
    SDEdit \cite{meng2022sdeditguidedimagesynthesis} & 4.50/4.75 & 2.00/\textbf{6.17} & 4.17/7.50 & 4.00/\textbf{7.50} & 4.67/\textbf{5.67} & 3.42/6.67 & 2.67/\underline{4.83} & 2.58/\textbf{7.50} & 5.25/6.83 & 4.83/\underline{8.00} & 3.81/\textbf{6.54} \\
    FlowEdit \cite{kulikov2024flowedit} & 7.67/\underline{5.00} & 5.45/\underline{5.18} & \underline{7.55}/6.91 & 6.40/5.80 & \underline{6.75}/4.83 & 7.00/\underline{6.75} & 6.25/4.50 & 5.09/6.18 & 7.08/6.50 & \underline{8.17}/\textbf{8.17} & 6.77/5.98 \\
    DynaEdit \cite{kulikov2026versatileeditingvideocontent} & \textbf{8.17}/4.83 & \textbf{5.64}/4.73 & 7.45/7.45 & \underline{6.91}/6.36 & \textbf{7.75}/5.00 & \underline{7.09}/6.27 & \underline{6.75}/4.50 & \underline{5.67}/\underline{6.25} & \underline{7.73}/6.55 & \textbf{8.55}/\underline{8.00} & \underline{7.17}/5.96 \\
    \recInvInv & 5.67/4.83 & 3.58/4.25 & 5.92/\textbf{7.83} & 4.92/\underline{6.83} & 5.50/4.25 & 4.67/6.67 & 4.17/4.17 & 3.67/5.83 & 6.08/\underline{7.33} & 5.17/6.50 & 4.93/5.85 \\
    \midrule
    \rowcolor{oursbg} \textcolor{ours}{\textbf{Ours}} & \underline{7.92}/\textbf{5.08} & \underline{5.50}/4.25 & \textbf{7.92}/\underline{7.67} & \textbf{7.00}/6.50 & \underline{6.75}/\underline{5.33} & \textbf{7.58}/\textbf{8.00} & \textbf{6.83}/\textbf{5.00} & \textbf{7.33}/5.00 & \textbf{8.00}/\textbf{7.58} & 8.00/6.67 & \textbf{7.28}/\underline{6.11} \\
    \bottomrule
  \end{tabular}
\end{table*}

\textbf{Additional quantitative results.}
\Cref{tab:pexels_v3_K4___NFE8_srct} complements the VLM-based evaluation of the main paper with classical metrics~\cite{ju2023direct}. Ours leads on PSNR, LPIPS, MSE and CLIP-BG. On text alignment, SDEdit scores highest but has the weakest source preservation, while Ours is second with the best preservation, at 6 NFE against 9--13 for the other source-preserving baselines.

\begin{figure*}[ht]
    \centering
    \includegraphics[width=1.0\linewidth]{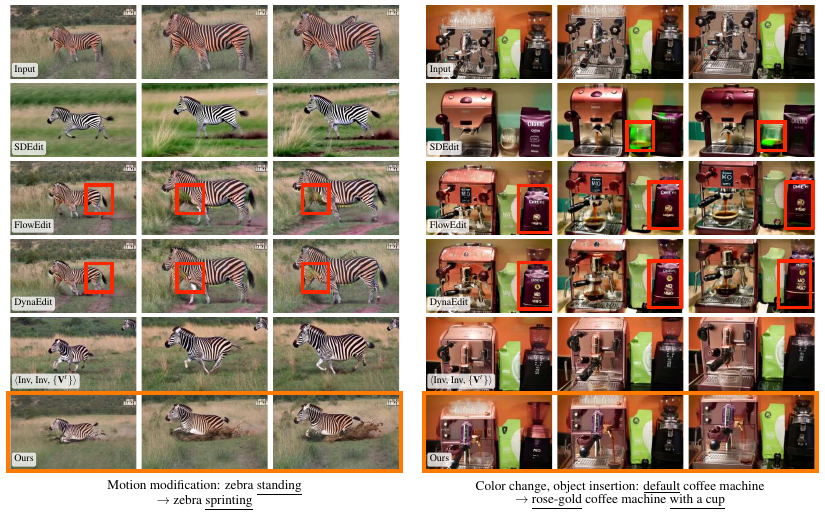}
    \caption{\textbf{Additional qualitative comparisons.} Red boxes highlight
    artifacts in baselines; our method (orange) preserves source appearance.
    Edited attributes are \underline{underlined}.}
    \label{fig:qualitative_appendix}
\end{figure*}

\textbf{Additional qualitative comparisons.}
\Cref{fig:qualitative_appendix} presents additional editing examples beyond
those in the main paper, covering further edit categories. The same pattern
holds: other methods exhibit burning artifacts (red boxes) or identity loss,
while our method (orange) preserves source appearance.

\begin{table*}[t]
\centering
\footnotesize
\caption{\textbf{Quantitative results with classical source-preservation metrics~\cite{ju2023direct} on our benchmark.} All methods run on the Reward Forcing distilled backbone with the default configurations of the main paper. Best in \textbf{bold}, second-best \underline{underlined}.}
\label{tab:pexels_v3_K4___NFE8_srct}
\setlength{\tabcolsep}{2.5pt}
\begin{tabular}{@{}l ccccc cc c@{}}
\toprule
& \multicolumn{5}{c}{\textbf{Source Preservation}} & \multicolumn{2}{c}{\textbf{Text Alignment}} & \\
\cmidrule(lr){2-6} \cmidrule(lr){7-8}
Method
& \shortstack{PSNR\\$\uparrow$} & \shortstack{SSIM\\$\times10^{2}$ $\uparrow$} & \shortstack{LPIPS\\$\times10^{3}$ $\downarrow$} & \shortstack{MSE\\$\times10^{4}$ $\downarrow$} & \shortstack{CLIP-BG\\$\times10^{2}$ $\uparrow$}
& \shortstack{CLIP-T\\$\times10^{2}$ $\uparrow$} & \shortstack{CLIP-T (edit)\\$\times10^{2}$ $\uparrow$}
& NFE $\downarrow$ \\
\midrule
SDEdit~\cite{meng2022sdeditguidedimagesynthesis}                 & 18.73 & 71.70 & 221.4 & 213.0 & 84.28 & \textbf{26.73} & \textbf{22.91} & \textbf{5} \\
FlowEdit~\cite{kulikov2024flowedit} & 24.00 & \underline{86.56} & 97.6 & 65.8 & 90.40 & 26.19 & 22.05 & 9 \\
DynaEdit~\cite{kulikov2026versatileeditingvideocontent}          & \underline{25.33} & \textbf{88.24} & \underline{83.6} & \underline{51.3} & \underline{91.41} & 25.91 & 21.91 & 13 \\
\recInvInv                                                   & 20.42 & 70.30 & 183.7 & 146.9 & 86.88 & 25.80 & 21.15 & 9 \\
\midrule
\rowcolor{oursbg}
\textcolor{ours}{\textbf{Ours}}                                  & \textbf{25.71} & 86.38 & \textbf{76.0} & \textbf{47.2} & \textbf{91.52} & \underline{26.21} & \underline{22.65} & \underline{6} \\
\bottomrule
\end{tabular}
\end{table*}

\begin{figure*}[ht]
    \centering
    \includegraphics[width=1.0\linewidth]{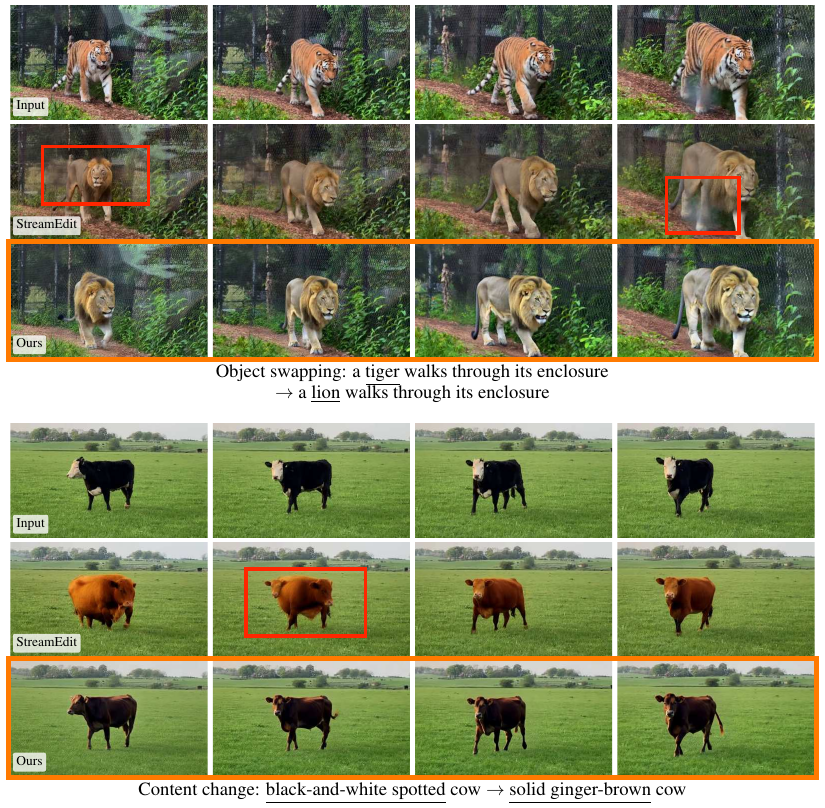}
 \caption{\textbf{Visual comparison with StreamEdit (4 editing steps) on Self Forcing.} Red boxes highlight artifacts; our method (orange) preserves the source appearance while applying the edit. Edited attributes in the prompts are \underline{underlined}.}   \label{fig:qualitative_appendix_on_self_forcing}
\end{figure*}

\textbf{Visual Comparison with StreamEdit.} \Cref{fig:qualitative_appendix_on_self_forcing} compares our method with StreamEdit on the Self Forcing backbone. StreamEdit renders the lion faintly against the scene and alters the cow's shape, whereas ours preserves the source layout and pose.

\textbf{V-cache scaling.} Table~\ref{tab:vcache_scaling} profiles the peak memory of our method across resolutions and chunk sizes on a single H100. Peak memory grows moderately along both axes. At the largest resolution tested ($960{\times}1664$), the bottleneck is the VAE decode of the larger chunk, which exceeds a single H100 at the default chunk size; reducing the chunk size to 1 restores feasibility. This also explains the slightly lower peak memory of methods on the bidirectional backbone in \Cref{tab:main_comparison}: their VAE encoding and decoding happen once for the full video, outside the denoising loop, rather than per chunk. We note that the DMD-distilled AR backbone is trained at $480{\times}832$, so generation quality outside this resolution is not guaranteed; the measurements characterize memory behavior only.

\begin{table}[t]
  \centering
  \small
  \caption{\textbf{Peak memory of our method across resolutions and chunk sizes.} Cells report peak memory (allocated~/~reserved, GB) on a single H100. OOM = out of memory.}
  \label{tab:vcache_scaling}
  \vspace{4pt}
  \begin{tabular}{@{}lcc@{}}
    \toprule
    Resolution (H$\times$W) & Chunk size 1 & Chunk size 3 (default) \\
    \midrule
    $240{\times}416$            & 19.9 / 21.3 & 20.7 / 22.4 \\
    $480{\times}832$ (default)  & 23.8 / 26.2 & 27.2 / 30.5 \\
    $960{\times}1664$           & 39.4 / 46.0 & OOM         \\
    \bottomrule
  \end{tabular}
\end{table}

%% file: sections/appendix/section_6.tex
\section{Additional Ablations}
\label{app:appendix_additional_abl}

All ablations in this section use the Reward Forcing backbone, our benchmark, and the metrics of \Cref{tab:main_comparison}. Every hyperparameter not under study is held at its default value ($S{=}S_{\mathrm{CtA}}{=}2$ with $K{=}4$, $b_{\min}{=}22$, $b_{\mathrm{CtA}}{=}19$, text-boost weight $4$).

\textbf{Reconstruction recipes under editing.} \Cref{tab:recipe_editing} evaluates every reconstruction recipe of \Cref{fig:main_figure} for editing on the Reward Forcing backbone. To isolate the effect of the recipe, we run Ours without commit-then-anchor and text boost. Two findings of the reconstruction study carry over to editing. First, the initial noise matters little once $\mathbf{V}$ comes from a clean forward pass: starting from random noise matches starting from inverted noise with the same $\mathbf{V}$, so the inversion passes can be dropped, reducing the cost from 10 to 6 NFE. Second, $\mathbf{V}$ must be extracted at low noise: moving the extraction timestep of the forward pass from $t{=}625$ to $t{=}1000$ lowers source adherence from 5.99 to 1.98 with random initialisation, and from 5.81 to 2.13 with inverted noise. Plain inversion (\emph{\recInv}) reaches the highest text adherence (6.88) only because it discards the source (2.58). Among the recipes that preserve the source, $\mathbf{V}$ from $t{=}625$ leads source adherence (5.99 vs.\ 5.73) at the same 6 NFE, while $t{=}0$ gives higher text adherence and TF. Overall, Ours (\emph{\recOurs}) offers the best trade-off: it runs at the lowest cost (6 NFE, against 9--10 for every recipe that relies on inversion), reaches the highest combined source and text adherence of all recipes, and attains the highest TF among all recipes at 6 NFE. Only \emph{\recRandInv} scores marginally higher on TF, at 9 NFE and with lower source adherence.

\begin{table*}[t]
  \centering
  \footnotesize
  \caption{\textbf{Editing quality across reconstruction recipes on the benchmark.} Best in \textbf{bold}, second-best \underline{underlined}. \emph{Ours} is reported without the commit-then-anchor schedule and text boost for a fair comparison.}
  \label{tab:recipe_editing}
  \setlength{\tabcolsep}{4pt}
  \begin{tabular}{@{}l ccccc c c@{}}
    \toprule
    & \multicolumn{2}{c}{\textbf{Editing Quality}} & \multicolumn{4}{c}{\textbf{Overall Visual Fidelity}} & \\
    \cmidrule(lr){2-3} \cmidrule(lr){4-7}
    Recipe
      & \shortstack{Source\\Adh.\,$\uparrow$} & \shortstack{Text\\Adh.\,$\uparrow$}
      & \shortstack{Video\\Quality\,$\uparrow$} & \shortstack{CLIP\\Frame\,$\uparrow$} & \shortstack{TF\,$\uparrow$} & \shortstack{TF$_{\text{u}}$\,$\uparrow$}
      & NFE\,$\downarrow$ \\
    \midrule
    \recInv & 2.58 & \textbf{6.88} & 7.54 & 0.781 & 0.257 & 0.200 & \underline{9} \\
    \recInvInv & 4.93 & \underline{5.85} & 7.93 & 0.847 & 0.383 & 0.323 & \underline{9} \\
    \midrule
    \recInvFwd[0] & 5.69 & 5.37 & 8.01 & 0.871 & 0.378 & 0.319 & 10 \\
    \recInvFwd[625] & \underline{5.81} & 5.12 & 8.05 & \underline{0.874} & 0.389 & 0.340 & 10 \\
    \recInvFwd[833] & 5.44 & 5.08 & \underline{8.07} & 0.867 & 0.370 & 0.311 & 10 \\
    \recInvFwd[938] & 3.52 & 4.83 & \textbf{8.08} & 0.837 & 0.356 & 0.281 & 10 \\
    \recInvFwd[1000] & 2.13 & 5.08 & 7.95 & 0.795 & 0.328 & 0.260 & 10 \\
    \midrule
    \recRandInv & 4.86 & 5.78 & 7.91 & 0.846 & \textbf{0.406} & \textbf{0.349} & \underline{9} \\
    \midrule
    \recRandFwd[625] & \textbf{5.99} & 5.09 & 8.04 & \textbf{0.876} & 0.393 & 0.332 & 6 \\
    \recRandFwd[833] & 5.36 & 4.99 & 8.06 & 0.866 & 0.388 & 0.333 & 6 \\
    \recRandFwd[938] & 3.63 & 5.12 & \underline{8.07} & 0.838 & 0.372 & 0.297 & 6 \\
    \recRandFwd[1000] & 1.98 & 5.13 & 7.96 & 0.794 & 0.329 & 0.261 & 6 \\
    \midrule
    \midrule
    \rowcolor{oursbg}
    \textcolor{ours}{\textbf{Ours: \recOurs}} & 5.73 & 5.42 & 7.97 & 0.871 & \underline{0.405} & \underline{0.342} & \textbf{6} \\
    \bottomrule
  \end{tabular}
\end{table*}

\begin{figure*}[ht]
    \centering
    \includegraphics[width=1.0\linewidth]{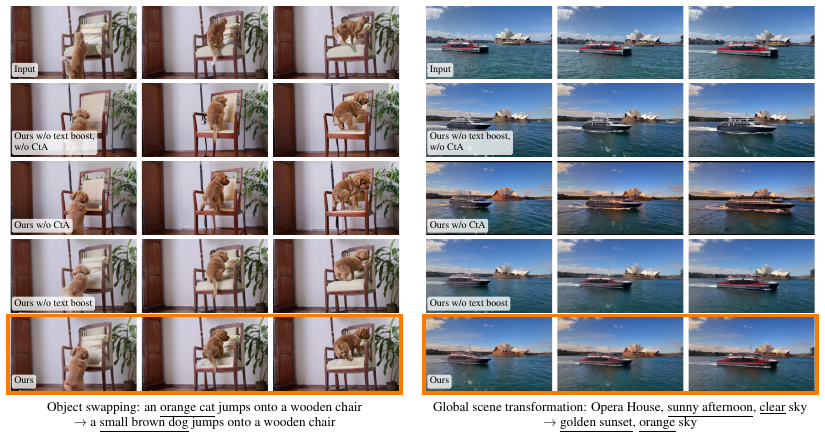}
\caption{\textbf{Effect of each component.} Without CtA the edit spreads beyond its target: the chair and the plant behind the cat drift, and the ferry changes from the source. Without text boost the edit is only partly applied: the animal keeps the cat's build and colour, and the sky keeps its blue. Ours swaps the cat for the dog and turns the sky golden while leaving chair, plant and ferry untouched.}

    \label{fig:method_components}
\end{figure*}

\textbf{Contribution of each component.} \Cref{tab:ablation_components} and \Cref{fig:method_components} start from the full configuration and disable commit-then-anchor (CtA) and text boost in turn. The two components act on different axes. CtA drives source preservation: added to the base configuration, it raises source adherence, CLIP Frame and TF$_{\text{u}}$, as the masked injection at the last steps keeps anchoring the unedited region to the source. Text boost drives the edit: alone, it raises text adherence but leaves source adherence unchanged and lowers CLIP Frame, since without CtA the last two steps run without any source $\mathbf{V}$. Combined, the full method keeps the source adherence of CtA, retains most of the text-adherence gain of the boost, and reaches the best TF. Video quality varies marginally across all four variants. \Cref{fig:method_components} shows the same pattern: without CtA, the sunset edit also alters the ferry, and without text boost, the sky keeps most of its blue.

\begin{table*}[t]
  \centering
  \footnotesize
\caption{\textbf{Component ablation on the benchmark.} Starting from the full \textbf{Ours} configuration, we disable each of the two components in turn: the commit-then-anchor schedule and text boost. Best in \textbf{bold}, second-best \underline{underlined}.}
  \label{tab:ablation_components}
  \setlength{\tabcolsep}{4pt}
  \begin{tabular}{@{}l cc cccc@{}}
    \toprule
    & \multicolumn{2}{c}{\textbf{Editing Quality}} & \multicolumn{4}{c}{\textbf{Overall Visual Fidelity}} \\
    \cmidrule(lr){2-3} \cmidrule(lr){4-7}
    Method
      & \shortstack{Source\\Adh.\,$\uparrow$} & \shortstack{Text\\Adh.\,$\uparrow$}
      & \shortstack{Video\\Quality\,$\uparrow$} & \shortstack{CLIP\\Frame\,$\uparrow$}
      & \shortstack{TF\,$\uparrow$} & \shortstack{TF$_{\text{u}}$\,$\uparrow$} \\
    \midrule
    Ours w/o commit-then-anchor, w/o text boost  & 5.73 & 5.42 & \underline{7.97} & 0.871 & 0.405 & 0.342 \\
    Ours w/o commit-then-anchor                  & 5.68 & \textbf{6.60} & \textbf{7.98} & 0.847 & 0.401 & 0.336 \\
    Ours w/o text boost                          & \underline{7.20} & 5.17 & \underline{7.97} & \textbf{0.900} & \underline{0.449} & \textbf{0.419} \\
    \rowcolor{oursbg}
    \textcolor{ours}{\textbf{Ours}}              & \textbf{7.28} & \underline{6.11} & 7.96 & \underline{0.885} & \textbf{0.473} & \underline{0.408} \\
    \bottomrule
  \end{tabular}
\end{table*}

\begin{figure*}[ht]
    \centering
    \includegraphics[width=1.0\linewidth]{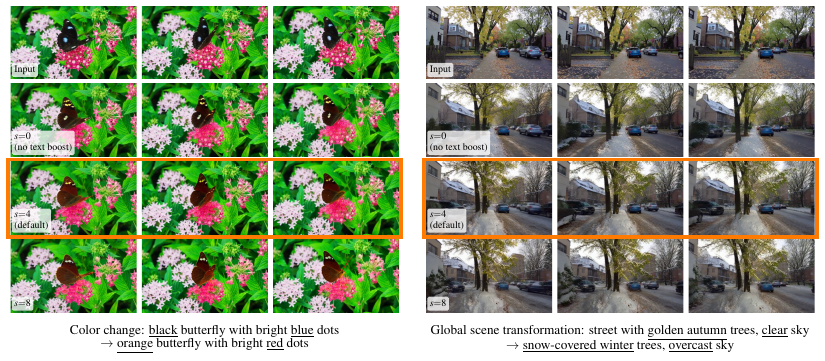}
\caption{\textbf{Effect of the text-boost weight $s$.} As $s$ grows, the butterfly turns more orange and the street more snow-covered: at $s{=}0$ the edit is incomplete, $s{=}4$ completes it, and $s{=}8$ adds little beyond it.}

    \label{fig:text_boost}
\end{figure*}

\textbf{Text-boost weight.} \Cref{tab:pexels_v3_ts_sweep} varies the text-boost weight $s$. Text adherence rises with $s$ but with diminishing returns: most of the gain, almost a full point, is reached by $s{=}4$, and larger weights add little. Beyond $s{=}4$, preservation degrades: source adherence and TF both peak at $s{=}4$ and decline for larger weights. We therefore set $s{=}4$. In \Cref{fig:text_boost}, the edit is only partially applied at $s{=}0$, complete at $s{=}4$, and changes little further at $s{=}8$.

\begin{table*}[t]
  \centering
  \footnotesize
 \caption{\textbf{Ablation on the weight $s$ of text-boost.} We ablate the text-boost weight from $0$ (no boost) to $8$. Best in \textbf{bold}, second-best \underline{underlined}. Our default is highlighted in orange.}
  \label{tab:pexels_v3_ts_sweep}
  \setlength{\tabcolsep}{4pt}
  \begin{tabular}{@{}l cc cccc@{}}
    \toprule
    & \multicolumn{2}{c}{\textbf{Editing Quality}} & \multicolumn{4}{c}{\textbf{Overall Visual Fidelity}} \\
    \cmidrule(lr){2-3} \cmidrule(lr){4-7}
    $s$
      & \shortstack{Source\\Adh.\,$\uparrow$} & \shortstack{Text\\Adh.\,$\uparrow$}
      & \shortstack{Video\\Quality\,$\uparrow$} & \shortstack{CLIP\\Frame\,$\uparrow$}
      & \shortstack{TF\,$\uparrow$} & \shortstack{TF$_{\text{u}}$\,$\uparrow$} \\
    \midrule
    0 & 7.20 & 5.17 & \textbf{7.97} & \textbf{0.900} & 0.449 & \underline{0.419} \\
    2 & \underline{7.25} & 5.60 & 7.96 & \underline{0.894} & \underline{0.465} & \textbf{0.427} \\
    \rowcolor{oursbg}
    \textcolor{ours}{\textbf{4}} & \textbf{7.28} & 6.11 & \underline{7.96} & 0.885 & \textbf{0.473} & 0.408 \\
    6 & 7.16 & \underline{6.21} & 7.92 & 0.876 & 0.433 & 0.403 \\
    8 & 7.11 & \textbf{6.27} & 7.88 & 0.868 & 0.438 & 0.401 \\
    \bottomrule
  \end{tabular}
\end{table*}

\begin{figure*}[ht]
    \centering
    \includegraphics[width=1.0\linewidth]{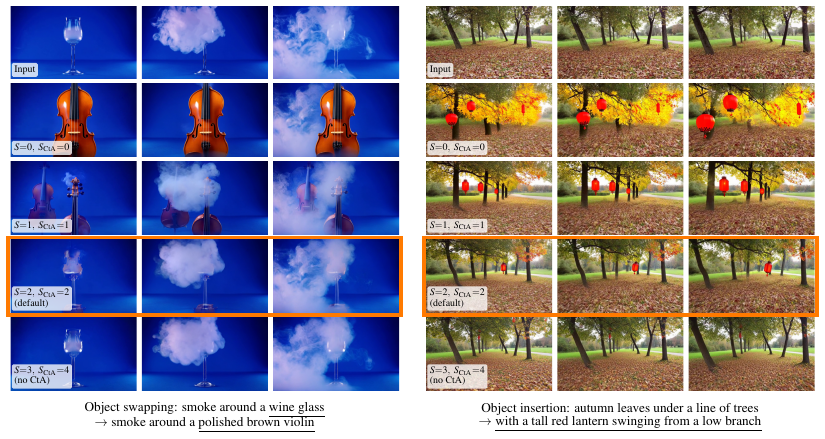}
\caption{\textbf{Effect of the injection schedule $(S, S_{\mathrm{CtA}})$.} Small $S$ re-renders the scene around the edit (a yellow canopy appears with the lantern); large $S$ reproduces the source and drops the edit (no lantern). The default $S{=}S_{\mathrm{CtA}}{=}2$ inserts the lantern and keeps the row of trees.}

    \label{fig:ablation_inj_step_appendix}
\end{figure*}

\textbf{Injection schedule.} \Cref{tab:abl_S_L} and \Cref{fig:ablation_inj_step_appendix} vary the number of injection steps $S$ and the step $S_{\mathrm{CtA}}$ at which masked injection starts, over the $K{=}4$ steps. Masked injection from the first step ($S{=}S_{\mathrm{CtA}}{=}0$) gives the highest text adherence but low source preservation and TF: at these steps the target sample is still close to noise, the mask read from its cross-attention marks no coherent region, and the layout forms without the source $\mathbf{V}$. One full injection step before the mask ($S{=}S_{\mathrm{CtA}}{=}1$) improves preservation only partly, and one full injection step followed by no injection ($S{=}1$, $S_{\mathrm{CtA}}{=}4$) drifts furthest from the source. At the other extreme, full injection over three of the four steps ($S{=}3$) preserves the source best but suppresses the edit. With $S{=}2$, the layout of the edit forms under full injection and the masked injection of the last two steps keeps the unedited region anchored: the default $S{=}S_{\mathrm{CtA}}{=}2$ reaches 7.28 source and 6.11 text adherence, with the highest TF among the schedules that do not suppress the edit. Steps left without injection between the two thresholds lower source adherence for every $S$ (at $S{=}2$: 7.28, 7.02 and 5.68 for $S_{\mathrm{CtA}}{=}2, 3, 4$), so we set $S_{\mathrm{CtA}}{=}S$. In the lantern example of \Cref{fig:ablation_inj_step_appendix}, $S{=}0$ re-renders the scene with a yellow canopy, $S{=}1$ inserts several lanterns and alters the background, $S{=}3$ reconstructs the source without the lantern, and the default inserts the lantern while keeping the row of trees from the source.

\begin{table*}[t]
  \centering
  \footnotesize
  \caption{\textbf{Ablation over the injection step ($S$) and the CtA start step ($S_{\mathrm{CtA}}$).} For each configuration we also list the per-step behaviour of the denoising steps: \textbf{full} = full $\mathbf{V}$-injection, \textbf{masked} = masked $\mathbf{V}$-injection (CtA anchor), \textbf{free} = no injection. $S_{\mathrm{CtA}}{=}4$ effectively disables CtA. Best per column in \textbf{bold}, second-best \underline{underlined}. Our default is highlighted in orange.}
  \label{tab:abl_S_L}
  \setlength{\tabcolsep}{3pt}
  \begin{tabular}{@{}cc cccc cc cccc@{}}
    \toprule
    \multicolumn{2}{c}{\textbf{Hyper-params}} & \multicolumn{4}{c}{\textbf{Per-step behaviour}} & \multicolumn{2}{c}{\textbf{Editing Quality}} & \multicolumn{4}{c}{\textbf{Overall Visual Fidelity}} \\
    \cmidrule(lr){1-2} \cmidrule(lr){3-6} \cmidrule(lr){7-8} \cmidrule(lr){9-12}
    $S$ & $S_{\mathrm{CtA}}$
      & Step 0 & Step 1 & Step 2 & Step 3
      & \shortstack{Source\\Adh.\,$\uparrow$} & \shortstack{Text\\Adh.\,$\uparrow$}
      & \shortstack{Video\\Quality\,$\uparrow$} & \shortstack{CLIP\\Frame\,$\uparrow$}
      & \shortstack{TF\,$\uparrow$} & \shortstack{TF$_{\text{u}}$\,$\uparrow$} \\
    \midrule
    0 & 0 & masked & masked & masked & masked & 5.81            & \textbf{7.47}    & 7.77            & 0.796            & 0.273            & 0.242            \\
    0 & 1 & free   & masked & masked & masked & 5.47            & 7.36             & 7.73            & 0.787            & 0.270            & 0.235            \\
    0 & 2 & free   & free   & masked & masked & 5.13            & 7.42             & 7.63            & 0.782            & 0.267            & 0.236            \\
    0 & 3 & free   & free   & free   & masked & 4.43            & 7.37             & 7.68            & 0.772            & 0.261            & 0.225            \\
    1 & 1 & full   & masked & masked & masked & 6.34            & 6.94             & 7.86            & 0.829            & 0.344            & 0.304            \\
    1 & 2 & full   & free   & masked & masked & 6.22            & 7.11             & 7.83            & 0.827            & 0.341            & 0.292            \\
    1 & 3 & full   & free   & free   & masked & 5.88            & 7.21             & 7.82            & 0.819            & 0.337            & 0.289            \\
    1 & 4 & full   & free   & free   & free   & 3.13            & \underline{7.43} & \textbf{8.04}   & 0.790            & 0.298            & 0.232            \\
     \rowcolor{oursbg} \textcolor{ours}{\textbf{2}} & \textcolor{ours}{\textbf{2}} & full   & full   & masked & masked & 7.28            & 6.11             & 7.96            & 0.885            & 0.473            & 0.408            \\
    2 & 3 & full   & full   & free   & masked & 7.02            & 6.13             & 7.93            & 0.879            & 0.439            & 0.395            \\
    2 & 4 & full   & full   & free   & free   & 5.68            & 6.60             & \underline{7.98}& 0.847            & 0.401            & 0.336            \\
    3 & 3 & full   & full   & full   & masked & \textbf{8.29}   & 4.19             & 7.92            & \textbf{0.943}   & \underline{0.549}& \textbf{0.522}   \\
    3 & 4 & full   & full   & full   & free   & \underline{7.73}& 4.33             & 7.90            & \underline{0.927}& \textbf{0.554}   & \underline{0.491}\\
    \bottomrule
  \end{tabular}
\end{table*}

\begin{figure*}[ht]
    \centering
    \includegraphics[width=1.0\linewidth]{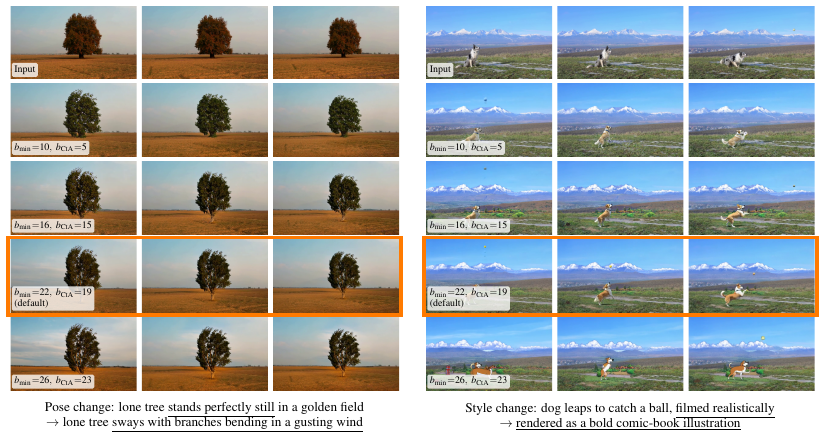}
    \caption{\textbf{Effect of the block thresholds $b_{\min}$ and $b_{\mathrm{CtA}}$.} Injecting the source $\mathbf{V}$ from early blocks on leaves the source nearly unchanged; injecting only in the last blocks strengthens the edit (the comic-book stylisation, the bending branches) at the cost of source appearance. The default $b_{\min}{=}22$, $b_{\mathrm{CtA}}{=}19$ sits between the two.}
    \label{fig:ablation_inj_block_appendix}
\end{figure*}

\textbf{Injection and mask block thresholds.} \Cref{tab:abl_mb_cta} and \Cref{fig:ablation_inj_block_appendix} vary the two block thresholds of the injection. $b_{\min}$ sets where the source $\mathbf{V}$ enters the network: in the blocks beyond $b_{\min}$, the target $\mathbf{V}$ is replaced by the source $\mathbf{V}$, in full during the first $S$ steps and outside the edit region, through the mask $M$, from step $S_{\mathrm{CtA}}$ on; the blocks up to $b_{\min}$ keep the target $\mathbf{V}$ at every step. $b_{\mathrm{CtA}}$ sets where $M$ comes from: the mask is the average of the cross-attention maps of the edit words over blocks $0$ to $b_{\mathrm{CtA}}$. Every configuration keeps $b_{\mathrm{CtA}} < b_{\min}$, so within a pass the mask is complete before the first injected block. $b_{\min}$ sets the balance between source preservation and edit. Replacing $\mathbf{V}$ from early blocks on ($b_{\min}{=}10$ or $16$) gives the highest source adherence, CLIP Frame and TF, but limits text adherence. Replacing $\mathbf{V}$ only in the last blocks ($b_{\min}{=}26$) raises text adherence, while source adherence and TF fall. Our default $b_{\min}{=}22$ lies between the two. For a fixed $b_{\min}$, averaging the mask over more blocks improves the edit. In \Cref{fig:ablation_inj_block_appendix}, the comic-book stylisation strengthens as $b_{\min}$ increases.

\begin{table*}[t]
  \centering
  \footnotesize
  \caption{\textbf{Ablation over the V-injection block $b_{\min}$ and the commit-then-anchor mask depth $b_{\mathrm{CtA}}$.} Best per column in \textbf{bold}, second-best \underline{underlined}. The orange row marks our default.}
  \label{tab:abl_mb_cta}
  \setlength{\tabcolsep}{4pt}
  \begin{tabular}{@{}cc cc cccc@{}}
    \toprule
    & & \multicolumn{2}{c}{\textbf{Editing Quality}} & \multicolumn{4}{c}{\textbf{Overall Visual Fidelity}} \\
    \cmidrule(lr){3-4} \cmidrule(lr){5-8}
    $b_{\min}$ & $b_{\mathrm{CtA}}$
      & \shortstack{Source\\Adh.\,$\uparrow$} & \shortstack{Text\\Adh.\,$\uparrow$}
      & \shortstack{Video\\Quality\,$\uparrow$} & \shortstack{CLIP\\Frame\,$\uparrow$}
      & \shortstack{TF\,$\uparrow$} & \shortstack{TF$_{\text{u}}$\,$\uparrow$} \\
    \midrule
    $10$ & $5$ & \underline{7.36} & 5.60 & 7.80 & \textbf{0.901} & \underline{0.481} & \textbf{0.463} \\
    $16$ & $5$ & 7.23 & 5.68 & 7.88 & \underline{0.899} & \textbf{0.482} & 0.427 \\
    $16$ & $10$ & \textbf{7.40} & 5.73 & 7.92 & 0.897 & \underline{0.481} & \underline{0.437} \\
    $16$ & $15$ & 7.32 & 5.81 & 7.92 & 0.897 & 0.479 & 0.428 \\
    $22$ & $5$ & 7.14 & 5.92 & 7.87 & 0.886 & 0.446 & 0.400 \\
    $22$ & $10$ & 7.33 & 6.02 & 7.91 & 0.883 & 0.449 & 0.409 \\
    $22$ & $15$ & 7.27 & 6.01 & 7.89 & 0.884 & 0.451 & 0.409 \\
    \rowcolor{oursbg}
    \textcolor{ours}{\textbf{22}} & \textcolor{ours}{\textbf{19}} & 7.28 & 6.11 & \underline{7.96} & 0.885 & 0.473 & 0.408 \\
    $26$ & $5$ & 6.65 & 6.65 & 7.93 & 0.860 & 0.414 & 0.395 \\
    $26$ & $10$ & 6.77 & 6.64 & 7.92 & 0.860 & 0.414 & 0.396 \\
    $26$ & $15$ & 6.74 & \textbf{6.70} & 7.94 & 0.861 & 0.415 & 0.392 \\
    $26$ & $19$ & 6.77 & \underline{6.67} & \underline{7.96} & 0.861 & 0.412 & 0.400 \\
    $26$ & $23$ & 6.77 & \underline{6.67} & \textbf{7.99} & 0.860 & 0.408 & 0.403 \\
    \bottomrule
  \end{tabular}
\end{table*}

%% file: sections/appendix/section_7.tex
\section{User Study Details}
\label{app:user_study_protocol}

\begin{figure*}[!t]
  \centering
  \includegraphics[width=0.95\linewidth]{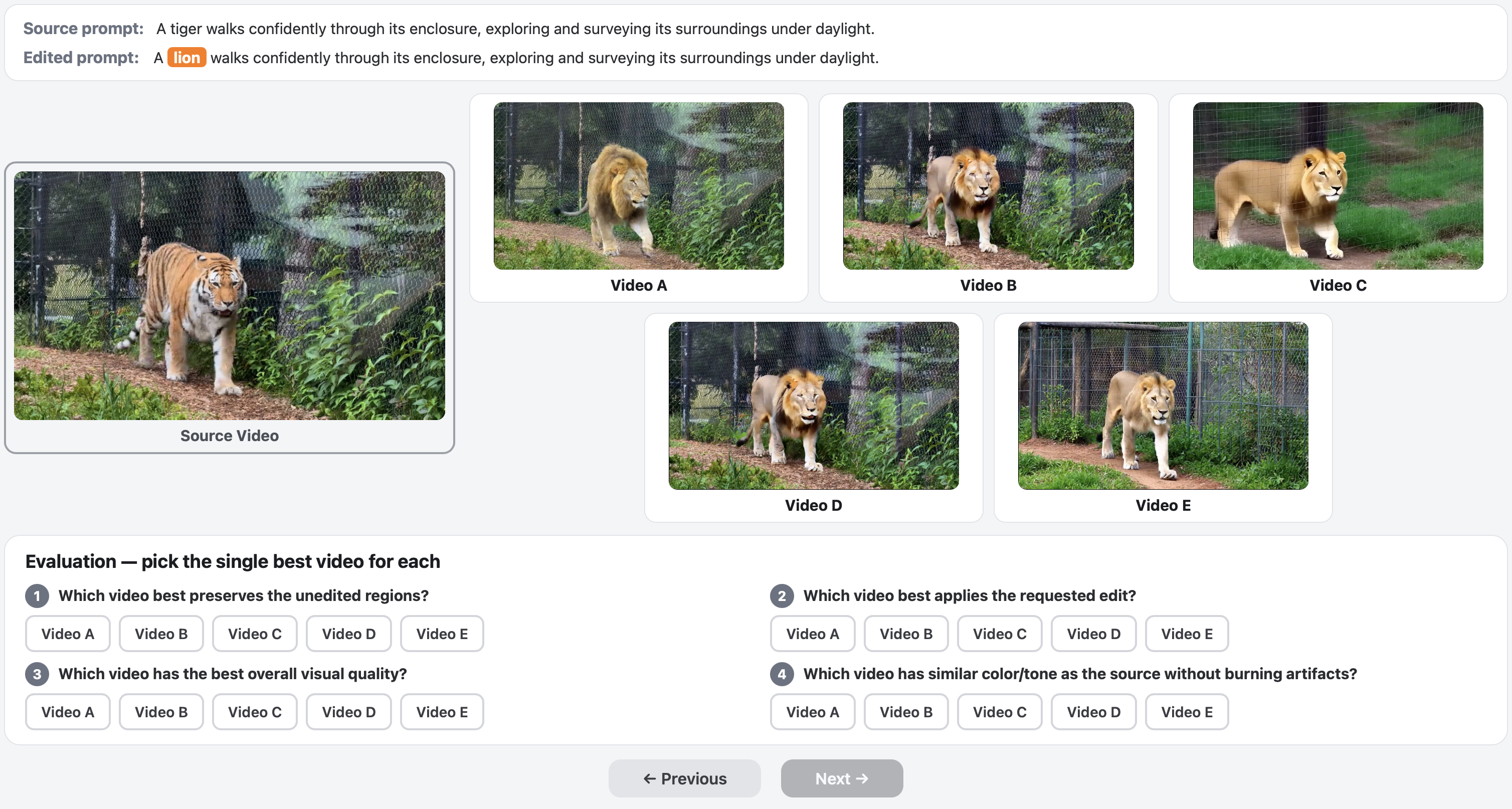}
\caption{\textbf{A screenshot from the user study set-up.}}
  \label{fig:user_study_screenshot}
\end{figure*}

Twenty volunteers evaluated 33 source videos, comparing our method against SDEdit, FlowEdit, DynaEdit, and \emph{\recInvInv} on the Reward Forcing backbone. For each video, participants viewed the source clip, the source and target prompts, and the five anonymized edited results in randomized order, and answered four forced-choice questions, one per criterion (source adherence, text adherence, visual quality, tonal fidelity), each time selecting the single best of the five results (\Cref{fig:user_study_screenshot}).

%% file: sections/appendix/section_5.tex
 \section{Limitations}
 \label{sec:limitations}

Our analysis of the QKV asymmetry covers four Wan~2.1-based models (Reward Forcing, CausVid, its bidirectional DMD-distilled variant, and the non-distilled teacher), all of which share the same DiT design, in which each block applies self-attention over video tokens followed by a separate cross-attention to the text conditioning; we have not verified the QKV asymmetry on architectures with different attention designs such as MM-DiT~\cite{esser2024scalingrectifiedflowtransformers}, which processes text and image tokens jointly in a single attention stream and where prior work has shown that V-injection can cause appearance leakage due to entangled text and image tokens~\cite{Shen_2025_ICCV}.

Because the $\mathbf{QKV}$ asymmetry is inherited from the bidirectional teacher (\Cref{fig:qkv_teacher}), $\mathbf{V}$-injection from a single clean pass may also apply to non-distilled bidirectional models. We do not evaluate this setting: bidirectional models cannot operate in the streaming regime we target, and there inversion is well-defined, so existing editors already inject source $\mathbf{V}$ from the inversion trajectory (\Cref{sec:prelim_inversion}). Whether our method offers advantages over those pipelines outside the streaming regime is left for future work.

Additionally, our method inherits the source--text trade-off common to training-free editors: on the Reward Forcing backbone, SDEdit attains the highest text adherence ($6.54$) by discarding the source ($3.81$), while FlowEdit and DynaEdit preserve the source at a cost in text adherence. Our method reaches a favourable balance between the two on distilled autoregressive video diffusion models with higher tonal fidelity, and exposes this balance through its hyperparameters: the injection budget $S$, the block threshold $b_{\min}$, and the text-boost scale. There is, however, no universal setting that works well for every example: more demanding edits may require per-example adjustment of these hyperparameters, and adaptive selection based on edit magnitude remains an open direction.

\section{Broader Societal Impact} \label{sec:app_broader_impact} 

Real-time video editing enables accessible creative tools for content creators, live broadcasting, and assistive applications, yet could also facilitate disinformation through manipulated live footage. We note that our method operates on an existing open-source backbone and does not introduce new generative capabilities; the same edits are achievable with the underlying model via slower pipelines. Nonetheless, reducing the barrier to real-time manipulation underscores the need for provenance tracking and watermarking in deployed video systems.   

\section{LLM Usage} 
\label{sec:llm_usage}  

We used Claude (Opus 4.6, Opus 5.5 and Fable 5.1)~\cite{anthropic2025claude} in three ways. First, to review drafts of the paper for grammar, clarity, phrasing and internal consistency; the model suggested sentence-level revisions, and the authors decided what to accept, reject or rephrase. Second, to assist with code development, including refactoring the implementation of AR-Edit for its public release; all generated code was reviewed and tested by the authors, and all experiments were run by the authors. Third, to write the plotting and drawing code for several figures and diagrams: we described the intended layout and content and iterated on the generated code until the output matched it, while the data shown and all design and presentation choices are the authors'. All scientific ideas, the method design and the analysis of results are the authors' own, and the authors take full responsibility for the content of this paper.

%% file: main.bib
@String(CVPR  = {IEEE Conf. Comput. Vis. Pattern Recog.})

@String(ICCV  = {Int. Conf. Comput. Vis.})

@String(ICLR  = {Int. Conf. Learn. Represent.})

@String(CVPR  = {CVPR})

@String(ICCV  = {ICCV})

@String(ICLR  = {ICLR})

@misc{esser2024scalingrectifiedflowtransformers,
      title={Scaling Rectified Flow Transformers for High-Resolution Image Synthesis}, 
      author={Patrick Esser and Sumith Kulal and Andreas Blattmann and Rahim Entezari and Jonas Müller and Harry Saini and Yam Levi and Dominik Lorenz and Axel Sauer and Frederic Boesel and Dustin Podell and Tim Dockhorn and Zion English and Kyle Lacey and Alex Goodwin and Yannik Marek and Robin Rombach},
      year={2024},
      eprint={2403.03206},
      archivePrefix={arXiv},
      primaryClass={cs.CV},
      url={https://arxiv.org/abs/2403.03206}, 
}

@misc{wan2025wanopenadvancedlargescale,
      title={Wan: Open and Advanced Large-Scale Video Generative Models}, 
      author={Team Wan and Ang Wang and Baole Ai and Bin Wen and Chaojie Mao and Chen-Wei Xie and Di Chen and Feiwu Yu and Haiming Zhao and Jianxiao Yang and Jianyuan Zeng and Jiayu Wang and Jingfeng Zhang and Jingren Zhou and Jinkai Wang and Jixuan Chen and Kai Zhu and Kang Zhao and Keyu Yan and Lianghua Huang and Mengyang Feng and Ningyi Zhang and Pandeng Li and Pingyu Wu and Ruihang Chu and Ruili Feng and Shiwei Zhang and Siyang Sun and Tao Fang and Tianxing Wang and Tianyi Gui and Tingyu Weng and Tong Shen and Wei Lin and Wei Wang and Wei Wang and Wenmeng Zhou and Wente Wang and Wenting Shen and Wenyuan Yu and Xianzhong Shi and Xiaoming Huang and Xin Xu and Yan Kou and Yangyu Lv and Yifei Li and Yijing Liu and Yiming Wang and Yingya Zhang and Yitong Huang and Yong Li and You Wu and Yu Liu and Yulin Pan and Yun Zheng and Yuntao Hong and Yupeng Shi and Yutong Feng and Zeyinzi Jiang and Zhen Han and Zhi-Fan Wu and Ziyu Liu},
      year={2025},
      eprint={2503.20314},
      archivePrefix={arXiv},
      primaryClass={cs.CV},
      url={https://arxiv.org/abs/2503.20314}, 
}

@misc{zheng2024opensorademocratizingefficientvideo,
      title={Open-Sora: Democratizing Efficient Video Production for All}, 
      author={Zangwei Zheng and Xiangyu Peng and Tianji Yang and Chenhui Shen and Shenggui Li and Hongxin Liu and Yukun Zhou and Tianyi Li and Yang You},
      year={2024},
      eprint={2412.20404},
      archivePrefix={arXiv},
      primaryClass={cs.CV},
      url={https://arxiv.org/abs/2412.20404}, 
}

@misc{kong2025hunyuanvideosystematicframeworklarge,
      title={HunyuanVideo: A Systematic Framework For Large Video Generative Models}, 
      author={Weijie Kong and Qi Tian and Zijian Zhang and Rox Min and Zuozhuo Dai and Jin Zhou and Jiangfeng Xiong and Xin Li and Bo Wu and Jianwei Zhang and Kathrina Wu and Qin Lin and Junkun Yuan and Yanxin Long and Aladdin Wang and Andong Wang and Changlin Li and Duojun Huang and Fang Yang and Hao Tan and Hongmei Wang and Jacob Song and Jiawang Bai and Jianbing Wu and Jinbao Xue and Joey Wang and Kai Wang and Mengyang Liu and Pengyu Li and Shuai Li and Weiyan Wang and Wenqing Yu and Xinchi Deng and Yang Li and Yi Chen and Yutao Cui and Yuanbo Peng and Zhentao Yu and Zhiyu He and Zhiyong Xu and Zixiang Zhou and Zunnan Xu and Yangyu Tao and Qinglin Lu and Songtao Liu and Dax Zhou and Hongfa Wang and Yong Yang and Di Wang and Yuhong Liu and Jie Jiang and Caesar Zhong},
      year={2025},
      eprint={2412.03603},
      archivePrefix={arXiv},
      primaryClass={cs.CV},
      url={https://arxiv.org/abs/2412.03603}, 
}

@misc{lipman2023flowmatchinggenerativemodeling,
      title={Flow Matching for Generative Modeling}, 
      author={Yaron Lipman and Ricky T. Q. Chen and Heli Ben-Hamu and Maximilian Nickel and Matt Le},
      year={2023},
      eprint={2210.02747},
      archivePrefix={arXiv},
      primaryClass={cs.LG},
      url={https://arxiv.org/abs/2210.02747}, 
}

@article{wang2024taming,
  title={Taming Rectified Flow for Inversion and Editing},
  author={Wang, Jiangshan and Pu, Junfu and Qi, Zhongang and Guo, Jiayi and Ma, Yue and Huang, Nisha and Chen, Yuxin and Li, Xiu and Shan, Ying},
  journal={arXiv preprint arXiv:2411.04746},
  year={2024}
}

@article{kulikov2024flowedit,
	title = {FlowEdit: Inversion-Free Text-Based Editing Using Pre-Trained Flow Models},
	author = {Kulikov, Vladimir and Kleiner, Matan and Huberman-Spiegelglas, Inbar and Michaeli, Tomer},
	journal = {arXiv preprint arXiv:2412.08629},
	year = {2024}
	}

@misc{kara2023raverandomizednoiseshuffling,
      title={RAVE: Randomized Noise Shuffling for Fast and Consistent Video Editing with Diffusion Models}, 
      author={Ozgur Kara and Bariscan Kurtkaya and Hidir Yesiltepe and James M. Rehg and Pinar Yanardag},
      year={2023},
      eprint={2312.04524},
      archivePrefix={arXiv},
      primaryClass={cs.CV},
      url={https://arxiv.org/abs/2312.04524}, 
}

@InProceedings{yin2025causvid,
    author    = {Yin, Tianwei and Zhang, Qiang and Zhang, Richard and Freeman, William T. and Durand, Fredo and Shechtman, Eli and Huang, Xun},
    title     = {From Slow Bidirectional to Fast Autoregressive Video Diffusion Models},
    booktitle = {Proceedings of the Computer Vision and Pattern Recognition Conference (CVPR)},
    month     = {June},
    year      = {2025},
    pages     = {22963-22974}
}

@misc{meng2022sdeditguidedimagesynthesis,
      title={SDEdit: Guided Image Synthesis and Editing with Stochastic Differential Equations}, 
      author={Chenlin Meng and Yutong He and Yang Song and Jiaming Song and Jiajun Wu and Jun-Yan Zhu and Stefano Ermon},
      year={2022},
      eprint={2108.01073},
      archivePrefix={arXiv},
      primaryClass={cs.CV},
      url={https://arxiv.org/abs/2108.01073}, 
}

@misc{liu2022flowstraightfastlearning,
      title={Flow Straight and Fast: Learning to Generate and Transfer Data with Rectified Flow}, 
      author={Xingchao Liu and Chengyue Gong and Qiang Liu},
      year={2022},
      eprint={2209.03003},
      archivePrefix={arXiv},
      primaryClass={cs.LG},
      url={https://arxiv.org/abs/2209.03003}, 
}

@misc{peebles2023scalablediffusionmodelstransformers,
      title={Scalable Diffusion Models with Transformers}, 
      author={William Peebles and Saining Xie},
      year={2023},
      eprint={2212.09748},
      archivePrefix={arXiv},
      primaryClass={cs.CV},
      url={https://arxiv.org/abs/2212.09748}, 
}

@misc{ma2025stepvideot2vtechnicalreportpractice,
      title={Step-Video-T2V Technical Report: The Practice, Challenges, and Future of Video Foundation Model}, 
      author={Guoqing Ma and Haoyang Huang and Kun Yan and Liangyu Chen and Nan Duan and Shengming Yin and Changyi Wan and Ranchen Ming and Xiaoniu Song and Xing Chen and Yu Zhou and Deshan Sun and Deyu Zhou and Jian Zhou and Kaijun Tan and Kang An and Mei Chen and Wei Ji and Qiling Wu and Wen Sun and Xin Han and Yanan Wei and Zheng Ge and Aojie Li and Bin Wang and Bizhu Huang and Bo Wang and Brian Li and Changxing Miao and Chen Xu and Chenfei Wu and Chenguang Yu and Dapeng Shi and Dingyuan Hu and Enle Liu and Gang Yu and Ge Yang and Guanzhe Huang and Gulin Yan and Haiyang Feng and Hao Nie and Haonan Jia and Hanpeng Hu and Hanqi Chen and Haolong Yan and Heng Wang and Hongcheng Guo and Huilin Xiong and Huixin Xiong and Jiahao Gong and Jianchang Wu and Jiaoren Wu and Jie Wu and Jie Yang and Jiashuai Liu and Jiashuo Li and Jingyang Zhang and Junjing Guo and Junzhe Lin and Kaixiang Li and Lei Liu and Lei Xia and Liang Zhao and Liguo Tan and Liwen Huang and Liying Shi and Ming Li and Mingliang Li and Muhua Cheng and Na Wang and Qiaohui Chen and Qinglin He and Qiuyan Liang and Quan Sun and Ran Sun and Rui Wang and Shaoliang Pang and Shiliang Yang and Sitong Liu and Siqi Liu and Shuli Gao and Tiancheng Cao and Tianyu Wang and Weipeng Ming and Wenqing He and Xu Zhao and Xuelin Zhang and Xianfang Zeng and Xiaojia Liu and Xuan Yang and Yaqi Dai and Yanbo Yu and Yang Li and Yineng Deng and Yingming Wang and Yilei Wang and Yuanwei Lu and Yu Chen and Yu Luo and Yuchu Luo and Yuhe Yin and Yuheng Feng and Yuxiang Yang and Zecheng Tang and Zekai Zhang and Zidong Yang and Binxing Jiao and Jiansheng Chen and Jing Li and Shuchang Zhou and Xiangyu Zhang and Xinhao Zhang and Yibo Zhu and Heung-Yeung Shum and Daxin Jiang},
      year={2025},
      eprint={2502.10248},
      archivePrefix={arXiv},
      primaryClass={cs.CV},
      url={https://arxiv.org/abs/2502.10248}, 
}

@misc{ho2022classifierfreediffusionguidance,
      title={Classifier-Free Diffusion Guidance}, 
      author={Jonathan Ho and Tim Salimans},
      year={2022},
      eprint={2207.12598},
      archivePrefix={arXiv},
      primaryClass={cs.LG},
      url={https://arxiv.org/abs/2207.12598}, 
}

@misc{albergo2023buildingnormalizingflowsstochastic,
      title={Building Normalizing Flows with Stochastic Interpolants}, 
      author={Michael S. Albergo and Eric Vanden-Eijnden},
      year={2023},
      eprint={2209.15571},
      archivePrefix={arXiv},
      primaryClass={cs.LG},
      url={https://arxiv.org/abs/2209.15571}, 
}

@misc{yin2024onestepdiffusiondistributionmatching,
      title={One-step Diffusion with Distribution Matching Distillation}, 
      author={Tianwei Yin and Michaël Gharbi and Richard Zhang and Eli Shechtman and Fredo Durand and William T. Freeman and Taesung Park},
      year={2024},
      eprint={2311.18828},
      archivePrefix={arXiv},
      primaryClass={cs.CV},
      url={https://arxiv.org/abs/2311.18828}, 
}

@misc{peng2025opensora20trainingcommerciallevel,
      title={Open-Sora 2.0: Training a Commercial-Level Video Generation Model in \$200k}, 
      author={Xiangyu Peng and Zangwei Zheng and Chenhui Shen and Tom Young and Xinying Guo and Binluo Wang and Hang Xu and Hongxin Liu and Mingyan Jiang and Wenjun Li and Yuhui Wang and Anbang Ye and Gang Ren and Qianran Ma and Wanying Liang and Xiang Lian and Xiwen Wu and Yuting Zhong and Zhuangyan Li and Chaoyu Gong and Guojun Lei and Leijun Cheng and Limin Zhang and Minghao Li and Ruijie Zhang and Silan Hu and Shijie Huang and Xiaokang Wang and Yuanheng Zhao and Yuqi Wang and Ziang Wei and Yang You},
      year={2025},
      eprint={2503.09642},
      archivePrefix={arXiv},
      primaryClass={cs.GR},
      url={https://arxiv.org/abs/2503.09642}, 
}

@misc{yin2024improveddistributionmatchingdistillation,
      title={Improved Distribution Matching Distillation for Fast Image Synthesis}, 
      author={Tianwei Yin and Michaël Gharbi and Taesung Park and Richard Zhang and Eli Shechtman and Fredo Durand and William T. Freeman},
      year={2024},
      eprint={2405.14867},
      archivePrefix={arXiv},
      primaryClass={cs.CV},
      url={https://arxiv.org/abs/2405.14867}, 
}

@article{ju2023direct,
  title={PnP Inversion: Boosting Diffusion-based Editing with 3 Lines of Code},
  author={Ju, Xuan and Zeng, Ailing and Bian, Yuxuan and Liu, Shaoteng and Xu, Qiang},
  journal={International Conference on Learning Representations ({ICLR})},
  year={2024}
}

@misc{anthropic2025claude,
  author = {Anthropic},
  title  = {Claude},
  year   = {2026},
  url    = {https://www.anthropic.com/claude}
}

@misc{rout2024semanticimageinversionediting,
      title={Semantic Image Inversion and Editing using Rectified Stochastic Differential Equations}, 
      author={Litu Rout and Yujia Chen and Nataniel Ruiz and Constantine Caramanis and Sanjay Shakkottai and Wen-Sheng Chu},
      year={2024},
      eprint={2410.10792},
      archivePrefix={arXiv},
      primaryClass={cs.LG},
      url={https://arxiv.org/abs/2410.10792}, 
}

@misc{huang2025selfforcingbridgingtraintest,
      title={Self Forcing: Bridging the Train-Test Gap in Autoregressive Video Diffusion}, 
      author={Xun Huang and Zhengqi Li and Guande He and Mingyuan Zhou and Eli Shechtman},
      year={2025},
      eprint={2506.08009},
      archivePrefix={arXiv},
      primaryClass={cs.CV},
      url={https://arxiv.org/abs/2506.08009}, 
}

@misc{liu2025rollingforcingautoregressivelong,
      title={Rolling Forcing: Autoregressive Long Video Diffusion in Real Time}, 
      author={Kunhao Liu and Wenbo Hu and Jiale Xu and Ying Shan and Shijian Lu},
      year={2025},
      eprint={2509.25161},
      archivePrefix={arXiv},
      primaryClass={cs.CV},
      url={https://arxiv.org/abs/2509.25161}, 
}

@misc{lu2025rewardforcingefficientstreaming,
      title={Reward Forcing: Efficient Streaming Video Generation with Rewarded Distribution Matching Distillation}, 
      author={Yunhong Lu and Yanhong Zeng and Haobo Li and Hao Ouyang and Qiuyu Wang and Ka Leong Cheng and Jiapeng Zhu and Hengyuan Cao and Zhipeng Zhang and Xing Zhu and Yujun Shen and Min Zhang},
      year={2025},
      eprint={2512.04678},
      archivePrefix={arXiv},
      primaryClass={cs.CV},
      url={https://arxiv.org/abs/2512.04678}, 
}

@misc{cui2025selfforcingminutescalehighqualityvideo,
      title={Self-Forcing++: Towards Minute-Scale High-Quality Video Generation}, 
      author={Justin Cui and Jie Wu and Ming Li and Tao Yang and Xiaojie Li and Rui Wang and Andrew Bai and Yuanhao Ban and Cho-Jui Hsieh},
      year={2025},
      eprint={2510.02283},
      archivePrefix={arXiv},
      primaryClass={cs.CV},
      url={https://arxiv.org/abs/2510.02283}, 
}

@misc{kodaira2026streamditrealtimestreamingtexttovideo,
      title={StreamDiT: Real-Time Streaming Text-to-Video Generation}, 
      author={Akio Kodaira and Tingbo Hou and Ji Hou and Markos Georgopoulos and Felix Juefei-Xu and Masayoshi Tomizuka and Yue Zhao},
      year={2026},
      eprint={2507.03745},
      archivePrefix={arXiv},
      primaryClass={cs.CV},
      url={https://arxiv.org/abs/2507.03745}, 
}

@misc{yu2025videossmautoregressivelongvideo,
      title={VideoSSM: Autoregressive Long Video Generation with Hybrid State-Space Memory}, 
      author={Yifei Yu and Xiaoshan Wu and Xinting Hu and Tao Hu and Yangtian Sun and Xiaoyang Lyu and Bo Wang and Lin Ma and Yuewen Ma and Zhongrui Wang and Xiaojuan Qi},
      year={2025},
      eprint={2512.04519},
      archivePrefix={arXiv},
      primaryClass={cs.CV},
      url={https://arxiv.org/abs/2512.04519}, 
}

@misc{hong2025sneakpeekfutureguidedinstructionalstreaming,
      title={SneakPeek: Future-Guided Instructional Streaming Video Generation}, 
      author={Cheeun Hong and German Barquero and Fadime Sener and Markos Georgopoulos and Edgar Schönfeld and Stefan Popov and Yuming Du and Oscar Mañas and Albert Pumarola},
      year={2025},
      eprint={2512.13019},
      archivePrefix={arXiv},
      primaryClass={cs.CV},
      url={https://arxiv.org/abs/2512.13019}, 
}

@misc{samuel2025lightningfastimageinversionediting,
      title={Lightning-Fast Image Inversion and Editing for Text-to-Image Diffusion Models}, 
      author={Dvir Samuel and Barak Meiri and Haggai Maron and Yoad Tewel and Nir Darshan and Shai Avidan and Gal Chechik and Rami Ben-Ari},
      year={2025},
      eprint={2312.12540},
      archivePrefix={arXiv},
      primaryClass={cs.CV},
      url={https://arxiv.org/abs/2312.12540}, 
}

@InProceedings{Shen_2025_ICCV,
    author    = {Shen, Tiancheng and Huang, Zilong and Li, Xiangtai and Lin, Zhijie and Liu, Jiyang and Wang, Yitong and Feng, Jiashi and Yang, Ming-Hsuan and Liew, Jun Hao},
    title     = {QK-Edit: Revisiting Attention-based Injection in MM-DiT for Image and Video Editing},
    booktitle = {Proceedings of the IEEE/CVF International Conference on Computer Vision (ICCV)},
    month     = {October},
    year      = {2025},
    pages     = {19043-19053}
}

@misc{dalva2024fluxspacedisentangledsemanticediting,
      title={FluxSpace: Disentangled Semantic Editing in Rectified Flow Transformers}, 
      author={Yusuf Dalva and Kavana Venkatesh and Pinar Yanardag},
      year={2024},
      eprint={2412.09611},
      archivePrefix={arXiv},
      primaryClass={cs.CV},
      url={https://arxiv.org/abs/2412.09611}, 
}

@misc{ouyang2025proeditinversionbasededitingprompts,
      title={ProEdit: Inversion-based Editing From Prompts Done Right}, 
      author={Zhi Ouyang and Dian Zheng and Xiao-Ming Wu and Jian-Jian Jiang and Kun-Yu Lin and Jingke Meng and Wei-Shi Zheng},
      year={2025},
      eprint={2512.22118},
      archivePrefix={arXiv},
      primaryClass={cs.CV},
      url={https://arxiv.org/abs/2512.22118}, 
}

@misc{cohen2024sliceditzeroshotvideoediting,
      title={Slicedit: Zero-Shot Video Editing With Text-to-Image Diffusion Models Using Spatio-Temporal Slices}, 
      author={Nathaniel Cohen and Vladimir Kulikov and Matan Kleiner and Inbar Huberman-Spiegelglas and Tomer Michaeli},
      year={2024},
      eprint={2405.12211},
      archivePrefix={arXiv},
      primaryClass={cs.CV},
      url={https://arxiv.org/abs/2405.12211}, 
}

@misc{tokenflow2023,
      title={TokenFlow: Consistent Diffusion Features for Consistent Video Editing}, 
      author={Michal Geyer and Omer Bar-Tal and Shai Bagon and Tali Dekel},
      year={2023},
      eprint={2307.10373},
      archivePrefix={arXiv},
      primaryClass={cs.CV},
      url={https://arxiv.org/abs/2307.10373}, 
}

@misc{zhang2025adaflowefficientlongvideo,
      title={AdaFlow: Efficient Long Video Editing via Adaptive Attention Slimming And Keyframe Selection}, 
      author={Shuheng Zhang and Yuqi Liu and Hongbo Zhou and Jun Peng and Yiyi Zhou and Xiaoshuai Sun and Rongrong Ji},
      year={2025},
      eprint={2502.05433},
      archivePrefix={arXiv},
      primaryClass={cs.CV},
      url={https://arxiv.org/abs/2502.05433}, 
}

@misc{yang2025videograinmodulatingspacetimeattention,
      title={VideoGrain: Modulating Space-Time Attention for Multi-grained Video Editing}, 
      author={Xiangpeng Yang and Linchao Zhu and Hehe Fan and Yi Yang},
      year={2025},
      eprint={2502.17258},
      archivePrefix={arXiv},
      primaryClass={cs.CV},
      url={https://arxiv.org/abs/2502.17258}, 
}

@misc{fan2024videoshoplocalizedsemanticvideo,
      title={Videoshop: Localized Semantic Video Editing with Noise-Extrapolated Diffusion Inversion}, 
      author={Xiang Fan and Anand Bhattad and Ranjay Krishna},
      year={2024},
      eprint={2403.14617},
      archivePrefix={arXiv},
      primaryClass={cs.CV},
      url={https://arxiv.org/abs/2403.14617}, 
}

@misc{wang2025videodirectorprecisevideoediting,
      title={VideoDirector: Precise Video Editing via Text-to-Video Models}, 
      author={Yukun Wang and Longguang Wang and Zhiyuan Ma and Qibin Hu and Kai Xu and Yulan Guo},
      year={2025},
      eprint={2411.17592},
      archivePrefix={arXiv},
      primaryClass={cs.CV},
      url={https://arxiv.org/abs/2411.17592}, 
}

@misc{bai2024unieditunifiedtuningfreeframework,
      title={UniEdit: A Unified Tuning-Free Framework for Video Motion and Appearance Editing}, 
      author={Jianhong Bai and Tianyu He and Yuchi Wang and Junliang Guo and Haoji Hu and Zuozhu Liu and Jiang Bian},
      year={2024},
      eprint={2402.13185},
      archivePrefix={arXiv},
      primaryClass={cs.CV},
      url={https://arxiv.org/abs/2402.13185}, 
}

@misc{yoon2025splitflowflowdecompositioninversionfree,
      title={SplitFlow: Flow Decomposition for Inversion-Free Text-to-Image Editing}, 
      author={Sung-Hoon Yoon and Minghan Li and Gaspard Beaudouin and Congcong Wen and Muhammad Rafay Azhar and Mengyu Wang},
      year={2025},
      eprint={2510.25970},
      archivePrefix={arXiv},
      primaryClass={cs.CV},
      url={https://arxiv.org/abs/2510.25970}, 
}

@misc{kim2025flowaligntrajectoryregularizedinversionfreeflowbased,
      title={FlowAlign: Trajectory-Regularized, Inversion-Free Flow-based Image Editing}, 
      author={Jeongsol Kim and Yeobin Hong and Jonghyun Park and Jong Chul Ye},
      year={2025},
      eprint={2505.23145},
      archivePrefix={arXiv},
      primaryClass={cs.CV},
      url={https://arxiv.org/abs/2505.23145}, 
}

@misc{li2026exacteditingflowbaseddiffusion,
      title={On Exact Editing of Flow-Based Diffusion Models}, 
      author={Zixiang Li and Yue Song and Jianing Peng and Ting Liu and Jun Huang and Xiaochao Qu and Luoqi Liu and Wei Wang and Yao Zhao and Yunchao Wei},
      year={2026},
      eprint={2512.24015},
      archivePrefix={arXiv},
      primaryClass={cs.CV},
      url={https://arxiv.org/abs/2512.24015}, 
}

@misc{jiao2025unieditflowunleashinginversionediting,
      title={UniEdit-Flow: Unleashing Inversion and Editing in the Era of Flow Models}, 
      author={Guanlong Jiao and Biqing Huang and Kuan-Chieh Wang and Renjie Liao},
      year={2025},
      eprint={2504.13109},
      archivePrefix={arXiv},
      primaryClass={cs.CV},
      url={https://arxiv.org/abs/2504.13109}, 
}

@misc{li2025flowdirectortrainingfreeflowsteering,
      title={FlowDirector: Training-Free Flow Steering for Precise Text-to-Video Editing}, 
      author={Guangzhao Li and Yanming Yang and Chenxi Song and Chi Zhang},
      year={2025},
      eprint={2506.05046},
      archivePrefix={arXiv},
      primaryClass={cs.CV},
      url={https://arxiv.org/abs/2506.05046}, 
}

@misc{kong2025tamingflowbasedi2vmodels,
      title={Taming Flow-based I2V Models for Creative Video Editing}, 
      author={Xianghao Kong and Hansheng Chen and Yuwei Guo and Lvmin Zhang and Gordon Wetzstein and Maneesh Agrawala and Anyi Rao},
      year={2025},
      eprint={2509.21917},
      archivePrefix={arXiv},
      primaryClass={cs.CV},
      url={https://arxiv.org/abs/2509.21917}, 
}

@misc{kulikov2026versatileeditingvideocontent,
      title={Versatile Editing of Video Content, Actions, and Dynamics without Training}, 
      author={Vladimir Kulikov and Roni Paiss and Andrey Voynov and Inbar Mosseri and Tali Dekel and Tomer Michaeli},
      year={2026},
      eprint={2603.17989},
      archivePrefix={arXiv},
      primaryClass={cs.CV},
      url={https://arxiv.org/abs/2603.17989}, 
}

@misc{lu2026chordeditonesteplowenergytransport,
      title={ChordEdit: One-Step Low-Energy Transport for Image Editing}, 
      author={Liangsi Lu and Xuhang Chen and Minzhe Guo and Shichu Li and Jingchao Wang and Yang Shi},
      year={2026},
      eprint={2602.19083},
      archivePrefix={arXiv},
      primaryClass={cs.CV},
      url={https://arxiv.org/abs/2602.19083}, 
}

@misc{feng2026streamdiffusionv2streamingdynamicinteractive,
      title={StreamDiffusionV2: A Streaming System for Dynamic and Interactive Video Generation}, 
      author={Tianrui Feng and Zhi Li and Shuo Yang and Haocheng Xi and Muyang Li and Xiuyu Li and Lvmin Zhang and Keting Yang and Kelly Peng and Song Han and Maneesh Agrawala and Kurt Keutzer and Akio Kodaira and Chenfeng Xu},
      year={2026},
      eprint={2511.07399},
      archivePrefix={arXiv},
      primaryClass={cs.CV},
      url={https://arxiv.org/abs/2511.07399}, 
}

@misc{bai2025qwen25vltechnicalreport,
      title={Qwen2.5-VL Technical Report}, 
      author={Shuai Bai and Keqin Chen and Xuejing Liu and Jialin Wang and Wenbin Ge and Sibo Song and Kai Dang and Peng Wang and Shijie Wang and Jun Tang and Humen Zhong and Yuanzhi Zhu and Mingkun Yang and Zhaohai Li and Jianqiang Wan and Pengfei Wang and Wei Ding and Zheren Fu and Yiheng Xu and Jiabo Ye and Xi Zhang and Tianbao Xie and Zesen Cheng and Hang Zhang and Zhibo Yang and Haiyang Xu and Junyang Lin},
      year={2025},
      eprint={2502.13923},
      archivePrefix={arXiv},
      primaryClass={cs.CV},
      url={https://arxiv.org/abs/2502.13923}, 
}

@misc{hsu2024autovfxphysicallyrealisticvideo,
      title={AutoVFX: Physically Realistic Video Editing from Natural Language Instructions}, 
      author={Hao-Yu Hsu and Zhi-Hao Lin and Albert Zhai and Hongchi Xia and Shenlong Wang},
      year={2024},
      eprint={2411.02394},
      archivePrefix={arXiv},
      primaryClass={cs.CV},
      url={https://arxiv.org/abs/2411.02394}, 
}

@misc{yatim2025dynvfxaugmentingrealvideos,
      title={DynVFX: Augmenting Real Videos with Dynamic Content}, 
      author={Danah Yatim and Rafail Fridman and Omer Bar-Tal and Tali Dekel},
      year={2025},
      eprint={2502.03621},
      archivePrefix={arXiv},
      primaryClass={cs.CV},
      url={https://arxiv.org/abs/2502.03621}, 
}

@misc{garibi2025tokenverseversatilemulticonceptpersonalization,
      title={TokenVerse: Versatile Multi-concept Personalization in Token Modulation Space}, 
      author={Daniel Garibi and Shahar Yadin and Roni Paiss and Omer Tov and Shiran Zada and Ariel Ephrat and Tomer Michaeli and Inbar Mosseri and Tali Dekel},
      year={2025},
      eprint={2501.12224},
      archivePrefix={arXiv},
      primaryClass={cs.CV},
      url={https://arxiv.org/abs/2501.12224}, 
}

@misc{peng2025dreambenchhumanalignedbenchmarkpersonalized,
      title={DreamBench++: A Human-Aligned Benchmark for Personalized Image Generation}, 
      author={Yuang Peng and Yuxin Cui and Haomiao Tang and Zekun Qi and Runpei Dong and Jing Bai and Chunrui Han and Zheng Ge and Xiangyu Zhang and Shu-Tao Xia},
      year={2025},
      eprint={2406.16855},
      archivePrefix={arXiv},
      primaryClass={cs.CV},
      url={https://arxiv.org/abs/2406.16855}, 
}

@misc{shin2026motionstreamrealtimevideogeneration,
      title={MotionStream: Real-Time Video Generation with Interactive Motion Controls}, 
      author={Joonghyuk Shin and Zhengqi Li and Richard Zhang and Jun-Yan Zhu and Jaesik Park and Eli Shechtman and Xun Huang},
      year={2026},
      eprint={2511.01266},
      archivePrefix={arXiv},
      primaryClass={cs.CV},
      url={https://arxiv.org/abs/2511.01266}, 
}

@misc{mokady2022nulltextinversioneditingreal,
      title={Null-text Inversion for Editing Real Images using Guided Diffusion Models}, 
      author={Ron Mokady and Amir Hertz and Kfir Aberman and Yael Pritch and Daniel Cohen-Or},
      year={2022},
      eprint={2211.09794},
      archivePrefix={arXiv},
      primaryClass={cs.CV},
      url={https://arxiv.org/abs/2211.09794}, 
}

@misc{song2022denoisingdiffusionimplicitmodels,
      title={Denoising Diffusion Implicit Models}, 
      author={Jiaming Song and Chenlin Meng and Stefano Ermon},
      year={2022},
      eprint={2010.02502},
      archivePrefix={arXiv},
      primaryClass={cs.LG},
      url={https://arxiv.org/abs/2010.02502}, 
}

@misc{deng2024fireflowfastinversionrectified,
      title={FireFlow: Fast Inversion of Rectified Flow for Image Semantic Editing}, 
      author={Yingying Deng and Xiangyu He and Changwang Mei and Peisong Wang and Fan Tang},
      year={2024},
      eprint={2412.07517},
      archivePrefix={arXiv},
      primaryClass={cs.CV},
      url={https://arxiv.org/abs/2412.07517}, 
}

@inproceedings{Avrahami_2025,
   title={Stable Flow: Vital Layers for Training-Free Image Editing},
   url={http://dx.doi.org/10.1109/CVPR52734.2025.00738},
   DOI={10.1109/cvpr52734.2025.00738},
   booktitle={2025 IEEE/CVF Conference on Computer Vision and Pattern Recognition (CVPR)},
   publisher={IEEE},
   author={Avrahami, Omri and Patashnik, Or and Fried, Ohad and Nemchinov, Egor and Aberman, Kfir and Lischinski, Dani and Cohen-Or, Daniel},
   year={2025},
   month=June, pages={7877–7888} }

@misc{su2023roformerenhancedtransformerrotary,
      title={RoFormer: Enhanced Transformer with Rotary Position Embedding}, 
      author={Jianlin Su and Yu Lu and Shengfeng Pan and Ahmed Murtadha and Bo Wen and Yunfeng Liu},
      year={2023},
      eprint={2104.09864},
      archivePrefix={arXiv},
      primaryClass={cs.CL},
      url={https://arxiv.org/abs/2104.09864}, 
}

@misc{radford2021learningtransferablevisualmodels,
      title={Learning Transferable Visual Models From Natural Language Supervision}, 
      author={Alec Radford and Jong Wook Kim and Chris Hallacy and Aditya Ramesh and Gabriel Goh and Sandhini Agarwal and Girish Sastry and Amanda Askell and Pamela Mishkin and Jack Clark and Gretchen Krueger and Ilya Sutskever},
      year={2021},
      eprint={2103.00020},
      archivePrefix={arXiv},
      primaryClass={cs.CV},
      url={https://arxiv.org/abs/2103.00020}, 
}

@inproceedings{vaswani2023attentionneed, author = {Vaswani, Ashish and Shazeer, Noam and Parmar, Niki and Uszkoreit, Jakob and Jones, Llion and Gomez, Aidan N. and Kaiser, {\L}ukasz and Polosukhin, Illia}, title = {Attention is all you need}, year = {2017}, isbn = {9781510860964}, publisher = {Curran Associates Inc.}, address = {Red Hook, NY, USA}, booktitle = {Proceedings of the 31st International Conference on Neural Information Processing Systems}, pages = {6000–6010}, numpages = {11}, location = {Long Beach, California, USA}, series = {NIPS'17} }

@misc{wu2024turboeditinstanttextbasedimage,
      title={TurboEdit: Instant text-based image editing}, 
      author={Zongze Wu and Nicholas Kolkin and Jonathan Brandt and Richard Zhang and Eli Shechtman},
      year={2024},
      eprint={2408.08332},
      archivePrefix={arXiv},
      primaryClass={cs.CV},
      url={https://arxiv.org/abs/2408.08332}, 
}

@misc{deutch2024turboedittextbasedimageediting,
      title={TurboEdit: Text-Based Image Editing Using Few-Step Diffusion Models}, 
      author={Gilad Deutch and Rinon Gal and Daniel Garibi and Or Patashnik and Daniel Cohen-Or},
      year={2024},
      eprint={2408.00735},
      archivePrefix={arXiv},
      primaryClass={cs.CV},
      url={https://arxiv.org/abs/2408.00735}, 
}

@misc{schuhmann2022iap,
  title        = {Improved Aesthetic Predictor},
  author       = {Schuhmann, Christoph},
  year         = {2022},
  howpublished = {\url{https://github.com/christophschuhmann/improved-aesthetic-predictor}},
  note         = {{GitHub} repository; MLP head fine-tuned on the SAC, LAION-Logos and AVA aesthetic ratings on top of {CLIP} {ViT-L/14} image features.}
}

@inproceedings{1763974.1764031, author = {Farneb{\"a}ck, Gunnar}, title = {Two-frame motion estimation based on polynomial expansion}, year = {2003}, isbn = {3540406018}, publisher = {Springer-Verlag}, address = {Berlin, Heidelberg}, booktitle = {Proceedings of the 13th Scandinavian Conference on Image Analysis}, pages = {363–370}, numpages = {8}, location = {Halmstad, Sweden}, series = {SCIA'03} }

@misc{pexels,
  author       = {{Pexels GmbH}},
  title        = {Pexels: Free stock photos, royalty free images \& videos},
  howpublished = {\url{https://www.pexels.com}},
  year         = {2026},
  note         = {Accessed: 2026-09-04}
}

@misc{kim2025reflextextguidededitingreal,
      title={ReFlex: Text-Guided Editing of Real Images in Rectified Flow via Mid-Step Feature Extraction and Attention Adaptation}, 
      author={Jimyeong Kim and Jungwon Park and Yeji Song and Nojun Kwak and Wonjong Rhee},
      year={2025},
      eprint={2507.01496},
      archivePrefix={arXiv},
      primaryClass={cs.CV},
      url={https://arxiv.org/abs/2507.01496}, 
}

@misc{cai2025ditctrlexploringattentioncontrol,
      title={DiTCtrl: Exploring Attention Control in Multi-Modal Diffusion Transformer for Tuning-Free Multi-Prompt Longer Video Generation}, 
      author={Minghong Cai and Xiaodong Cun and Xiaoyu Li and Wenze Liu and Zhaoyang Zhang and Yong Zhang and Ying Shan and Xiangyu Yue},
      year={2025},
      eprint={2412.18597},
      archivePrefix={arXiv},
      primaryClass={cs.CV},
      url={https://arxiv.org/abs/2412.18597}, 
}

@misc{zhang2018unreasonableeffectivenessdeepfeatures,
      title={The Unreasonable Effectiveness of Deep Features as a Perceptual Metric}, 
      author={Richard Zhang and Phillip Isola and Alexei A. Efros and Eli Shechtman and Oliver Wang},
      year={2018},
      eprint={1801.03924},
      archivePrefix={arXiv},
      primaryClass={cs.CV},
      url={https://arxiv.org/abs/1801.03924}, 
}

@misc{jiao2026streamedittrainingfreevideoediting,
      title={StreamEdit: Training-Free Video Editing via Few-Step Streaming Video Generation}, 
      author={Guanlong Jiao and Chenyangguang Zhang and Jia Jun Cheng Xian and Zewei Zhang and Renjie Liao},
      year={2026},
      eprint={2605.21466},
      archivePrefix={arXiv},
      primaryClass={cs.CV},
      url={https://arxiv.org/abs/2605.21466}, 
}

@article{needleman1970general,
  author = {Needleman, Saul B. and Wunsch, Christian D.},
  description = {This article describes the iterative matrix algorithm created by Needleman and Wunsch to analyze and compare amino acid sequences. They wanted to create a computer-operable algorithm to replace the visual comparisons used to study proteins at the time. This algorithm proved so useful it is still in use, with some modifications, today, and has inspired other algorithms, such as the Smith-Waterman algorithm, as well.},
  doi = {10.1016/0022-2836(70)90057-4},
  journal = {Journal of Molecular Biology},
  month = mar,
  number = 3,
  pages = {443-453},
  title = {A General Method Applicable to the Search for Similarities in the Amino Acid Sequence of Two Proteins},
  volume = 48,
  year = 1970
}

@misc{li2025egoeditdatasetrealtimestreaming,
      title={EgoEdit: Dataset, Real-Time Streaming Model, and Benchmark for Egocentric Video Editing}, 
      author={Runjia Li and Moayed Haji-Ali and Ashkan Mirzaei and Chaoyang Wang and Arpit Sahni and Ivan Skorokhodov and Aliaksandr Siarohin and Tomas Jakab and Junlin Han and Sergey Tulyakov and Philip Torr and Willi Menapace},
      year={2025},
      eprint={2512.06065},
      archivePrefix={arXiv},
      primaryClass={cs.CV},
      url={https://arxiv.org/abs/2512.06065}, 
}
